\def\year{2021}\relax
\documentclass[letterpaper]{article} 
\usepackage{aaai21}  
\usepackage{times}  
\usepackage{helvet} 
\usepackage{courier}  
\usepackage[hyphens]{url}  
\usepackage{graphicx} 
\usepackage{natbib}  
\usepackage{caption} 
\usepackage[utf8]{inputenc}
\usepackage{nicefrac}
\usepackage{latexsym}
\usepackage{microtype}
\usepackage{amsmath}
\usepackage{amssymb}
\usepackage{booktabs}
\usepackage{xcolor}
\usepackage{bm}
\usepackage{etoolbox}
\usepackage{algorithm2e}
\usepackage{threeparttable}
\usepackage{subcaption}

\newcommand{\tb}[1]{\textbf{#1}}
\newcommand{\vect}[1]{\boldsymbol{\mathbf{#1}}}

\title{Meta Label Correction for Noisy Label Learning}

\author{
    Guoqing Zheng,
    Ahmed Hassan Awadallah,
    Susan Dumais\\
}

\affiliations{
    Microsoft Research\\
    \{zheng, hassanam, sdumais\}@microsoft.com
}

\begin{document}
\maketitle

\begin{abstract}

Leveraging weak or noisy supervision for building effective machine
learning models has long been an important research problem. Its
importance has further increased recently due to the growing need for
large-scale datasets to train deep learning models. Weak or noisy
supervision could originate from multiple sources including non-expert
annotators or automatic labeling based on heuristics or user
interaction signals. There is an extensive amount of previous work
focusing on leveraging noisy labels. Most notably, recent work has
shown impressive gains by using a meta-learned instance re-weighting
approach where a meta-learning framework is used to assign instance
weights to noisy labels. In this paper, we extend this approach via
posing the problem as a label correction problem within a
meta-learning framework. We view the label correction procedure as a
meta-process and propose a new meta-learning based framework termed
MLC (Meta Label Correction) for learning with noisy
labels. Specifically, a label correction network is adopted as a
meta-model to produce corrected labels for noisy labels while the main
model is trained to leverage the corrected labels. Both models are
jointly trained by solving a bi-level optimization problem. We run
extensive experiments with different label noise levels and types on
both image recognition and text classification tasks. We compare the
re-weighing and correction approaches showing that the correction
framing addresses some of the limitations of re-weighting.  We also
show that the proposed MLC approach outperforms previous methods in
both image and language tasks.
\end{abstract}

\section{Introduction}

Recent advances in deep learning have enabled impressive performance on various tasks, including image
recognition~\cite{he2016deep} and natural language
processing~\cite{devlin2018bert}.
At the core of this success lies the availability of large amounts of
annotated data. However, such datasets are not readily available at
scale for many tasks. Learning with weak supervision aims to address
this challenge by leveraging weak evidences of supervision. Weak
supervision can come in several forms including: incomplete
supervision; where only a small subset of the training data has
labels, inexact supervision; where only coarse-grained annotations are
available, and inaccurate supervision; where noisy labels are given
~\cite{zhou2017brief}. In this work, we focus on using inaccurate
(noisy) labels as a form of weak supervision. Noisy labels may originate from multiple sources including: corrupted labels, non-expert annotators, automatic labels based on heuristics or user interaction signals, etc.

\begin{figure}[t]
  \centering
  \includegraphics[width=0.32\linewidth]{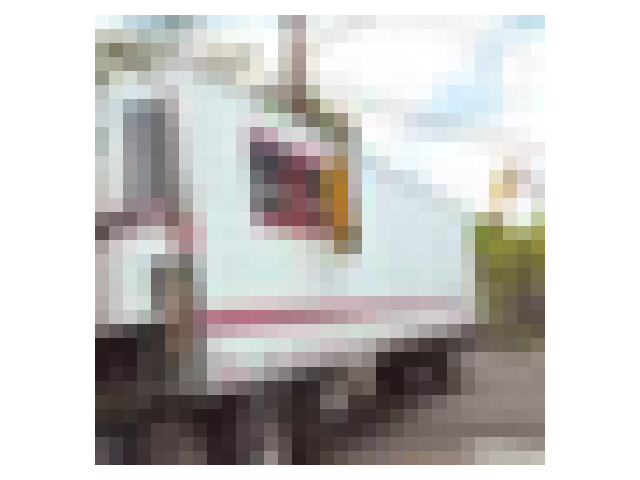}
  \includegraphics[width=0.32\linewidth]{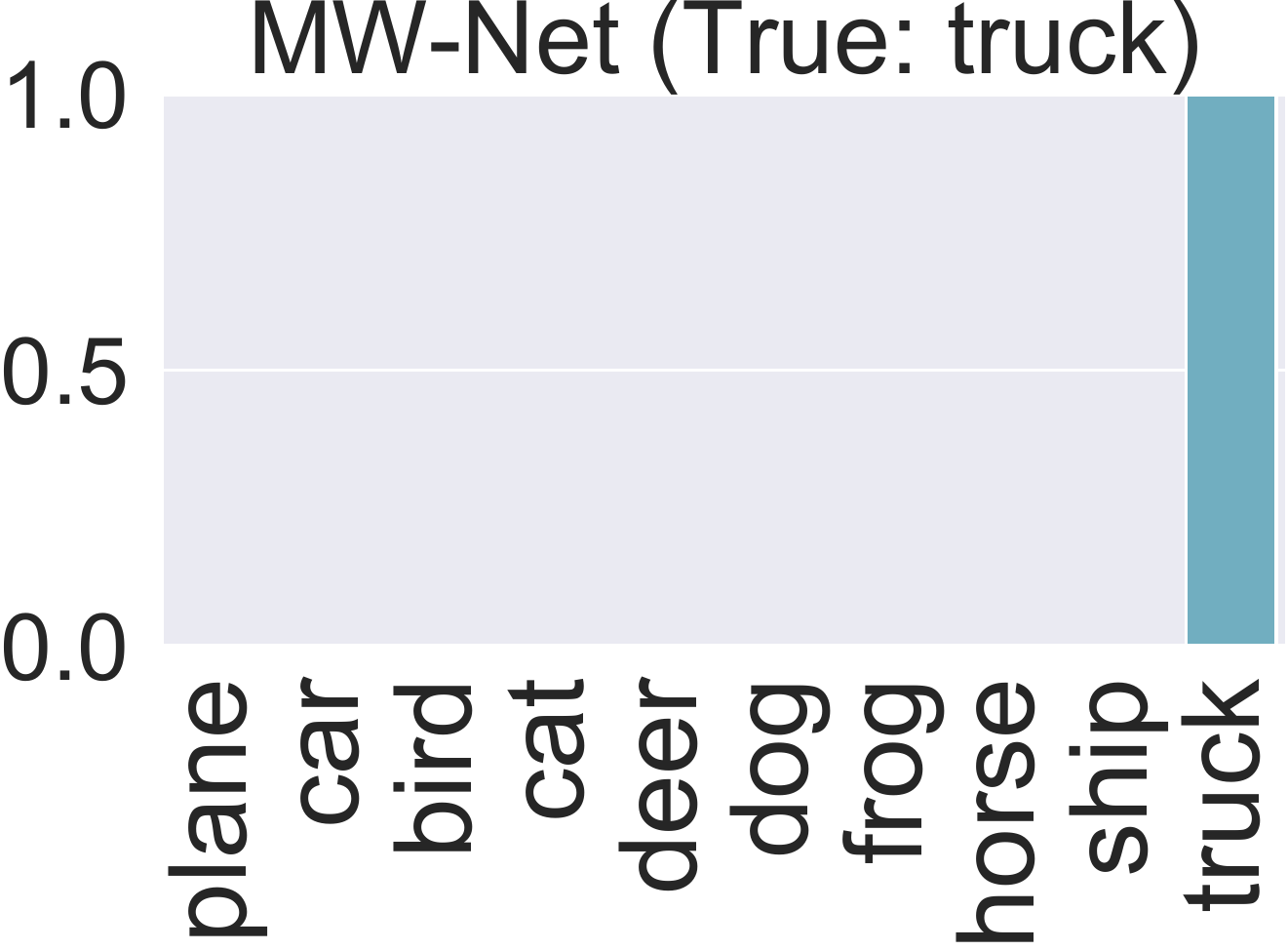}
  \includegraphics[width=0.32\linewidth]{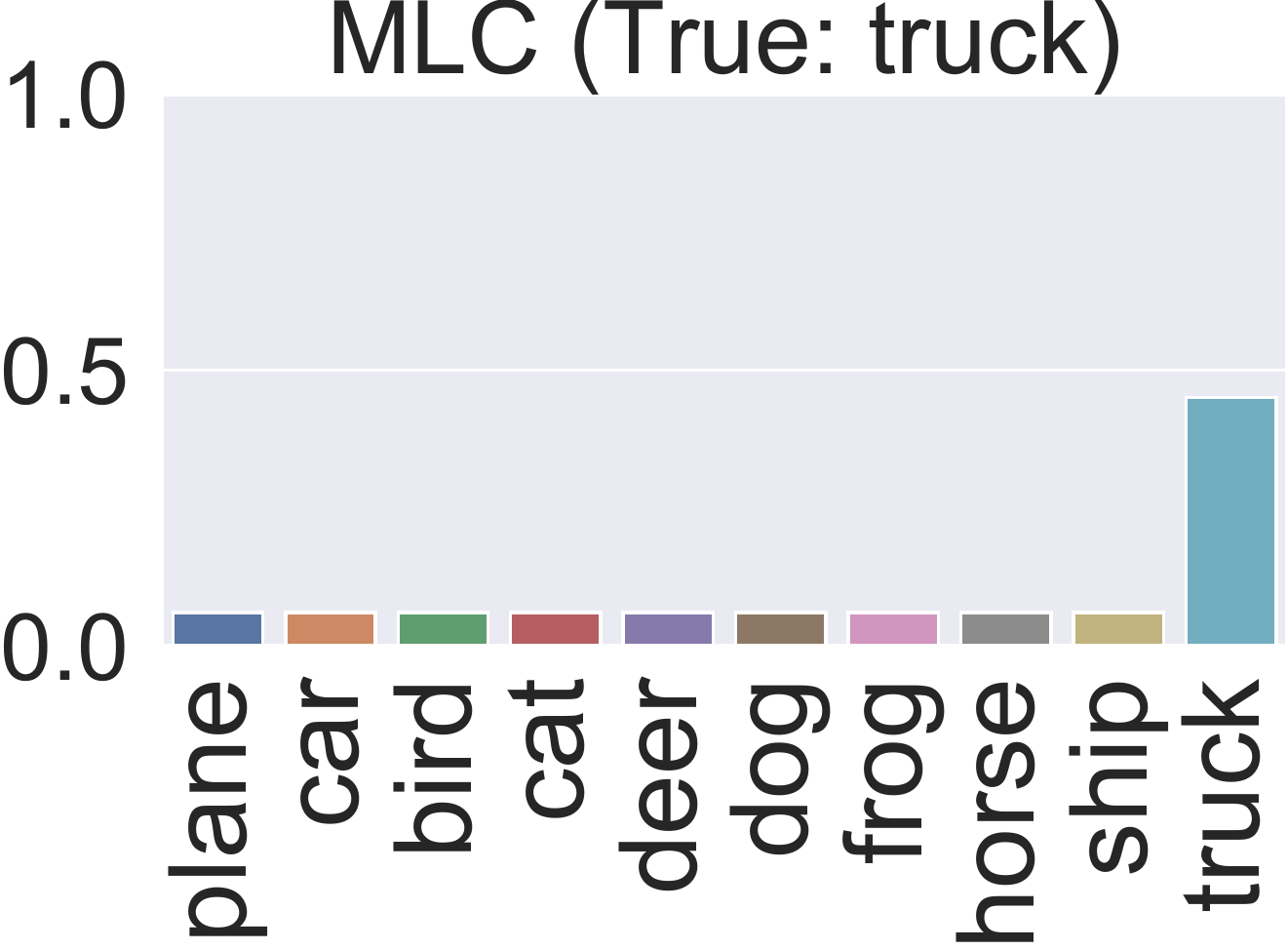}
  \includegraphics[width=0.32\linewidth]{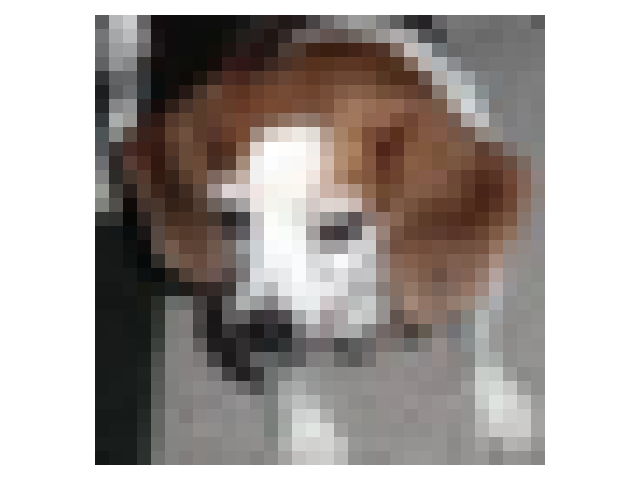}
  \includegraphics[width=0.32\linewidth]{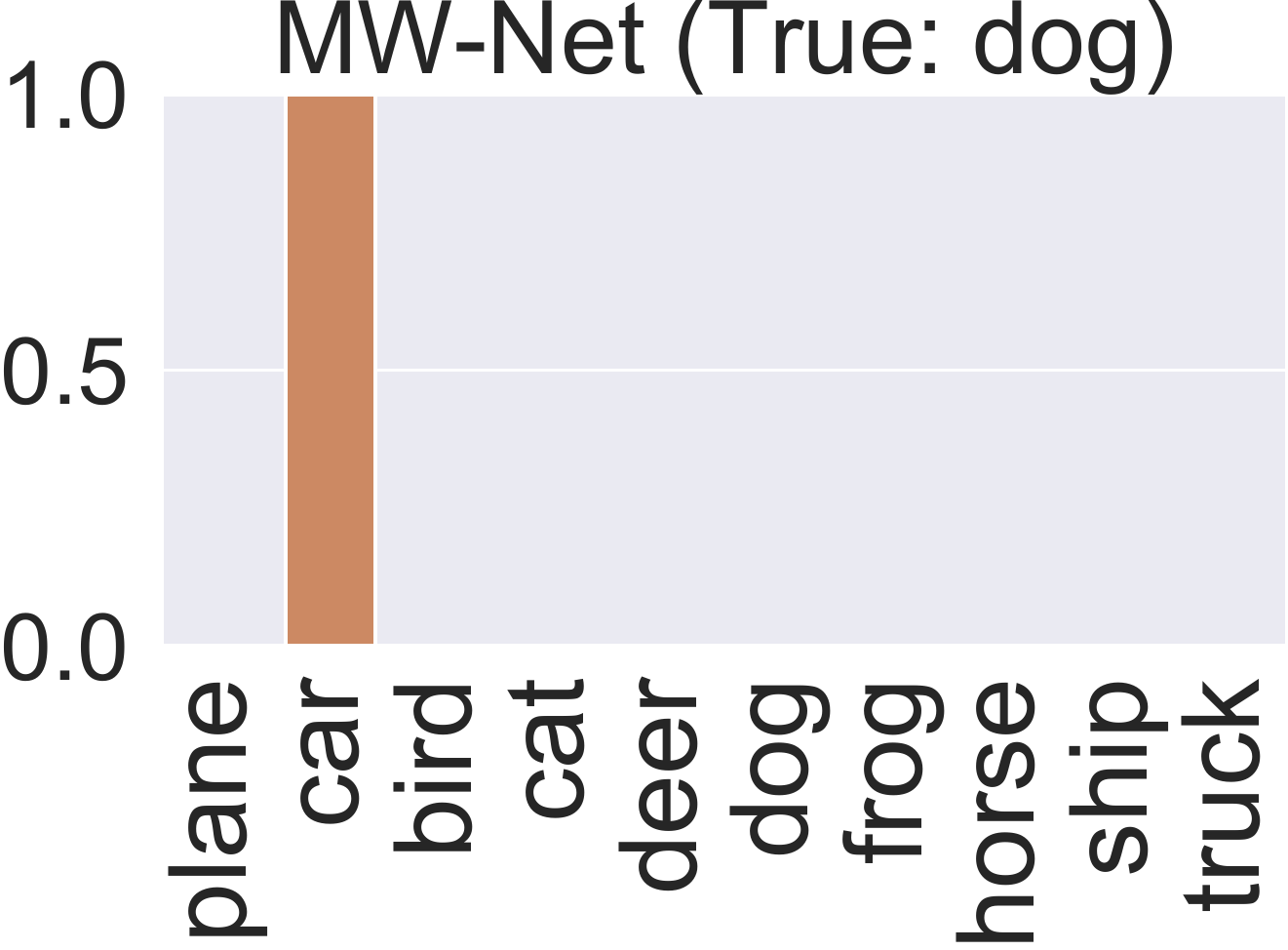}
  \includegraphics[width=0.32\linewidth]{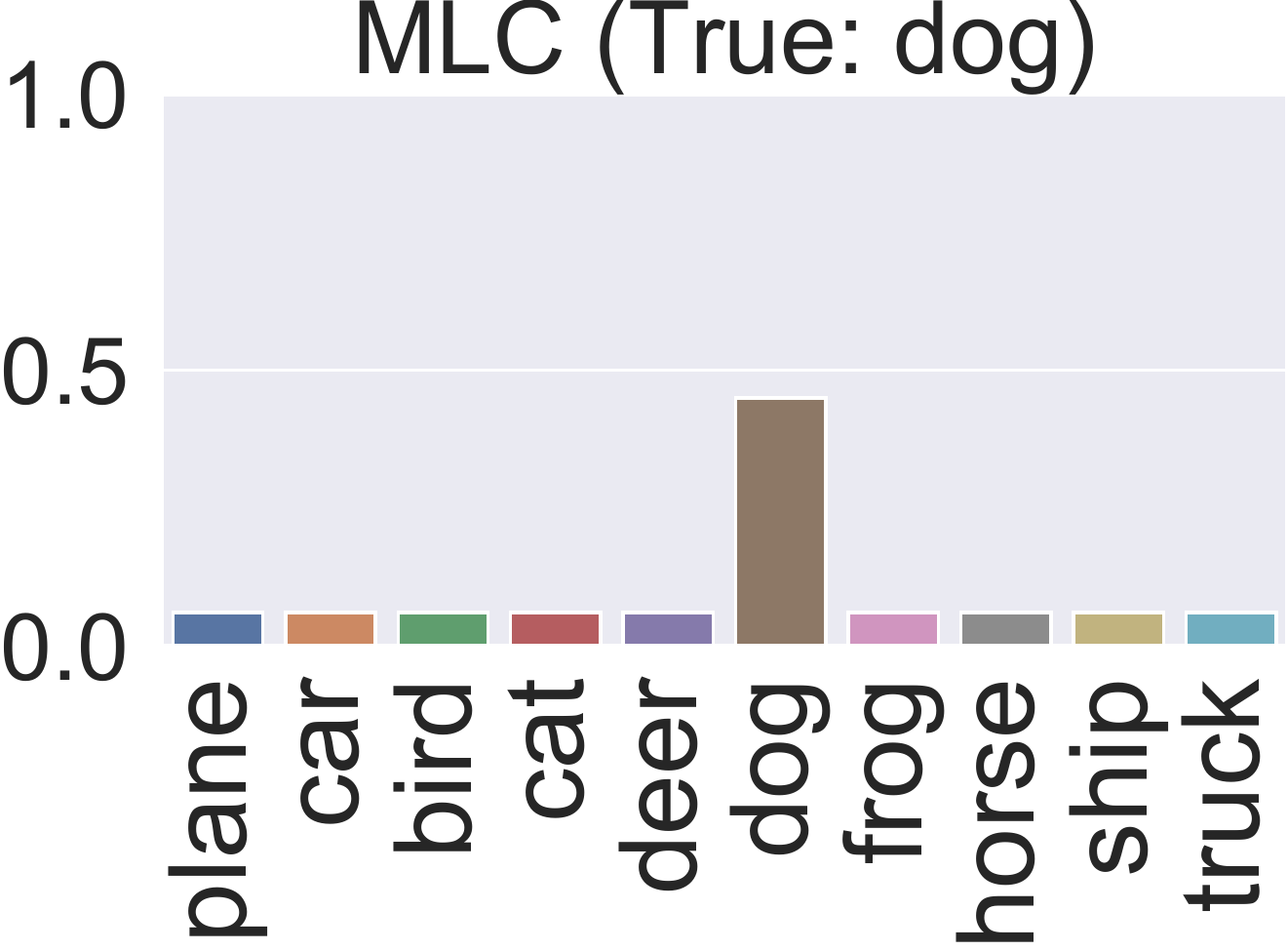}
  \includegraphics[width=0.32\linewidth]{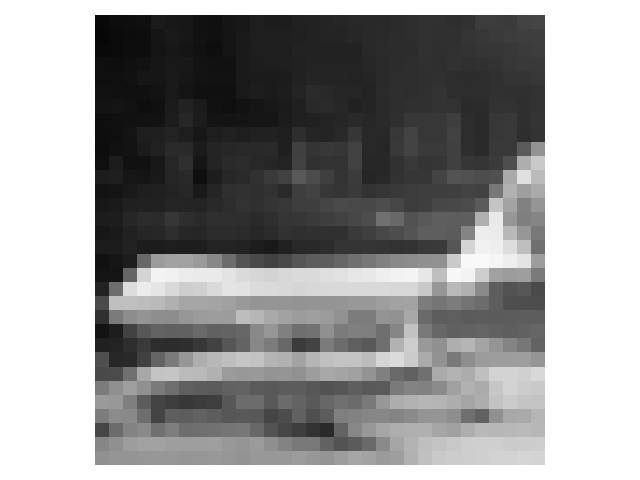}
  \includegraphics[width=0.32\linewidth]{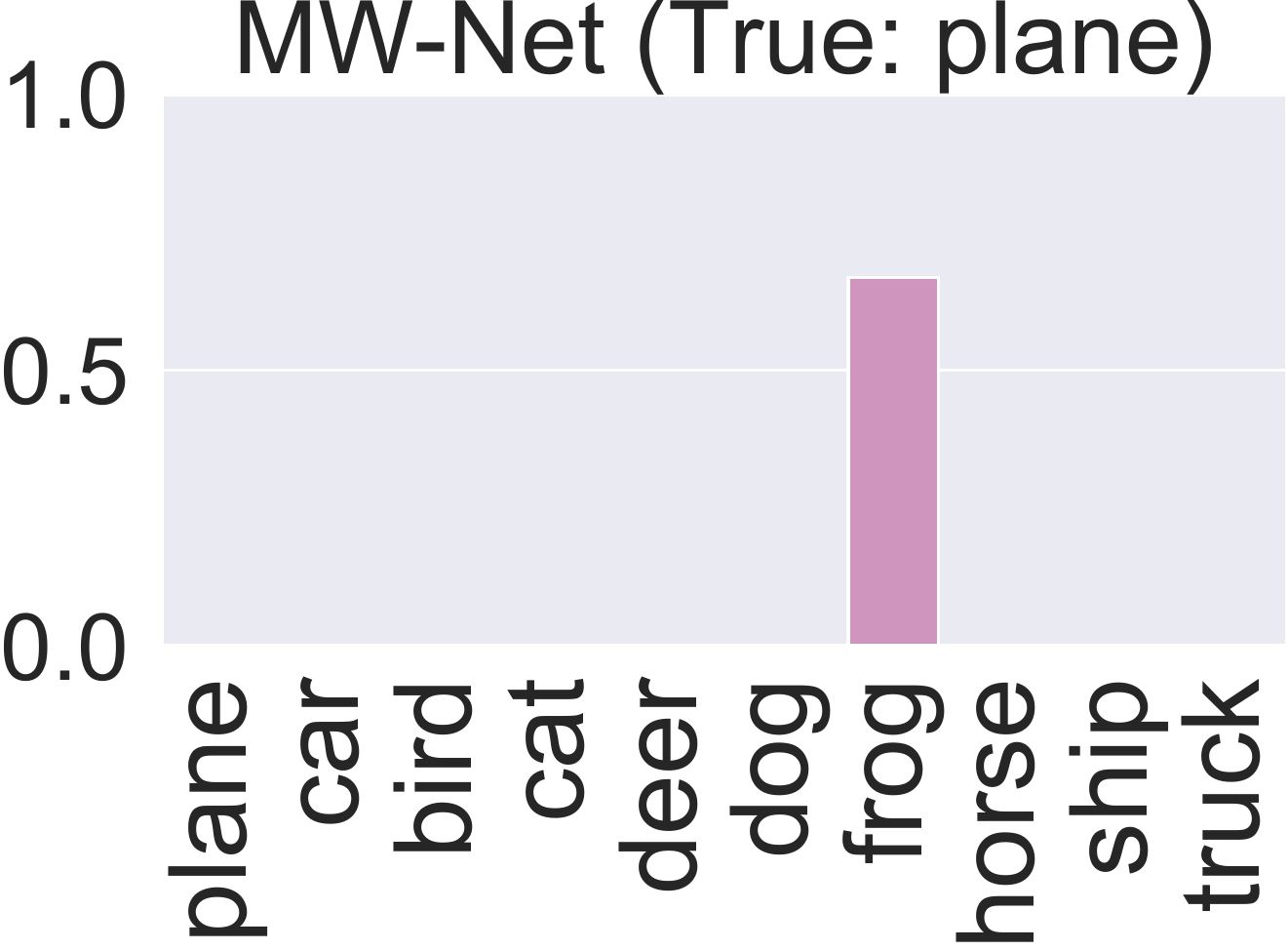}
  \includegraphics[width=0.32\linewidth]{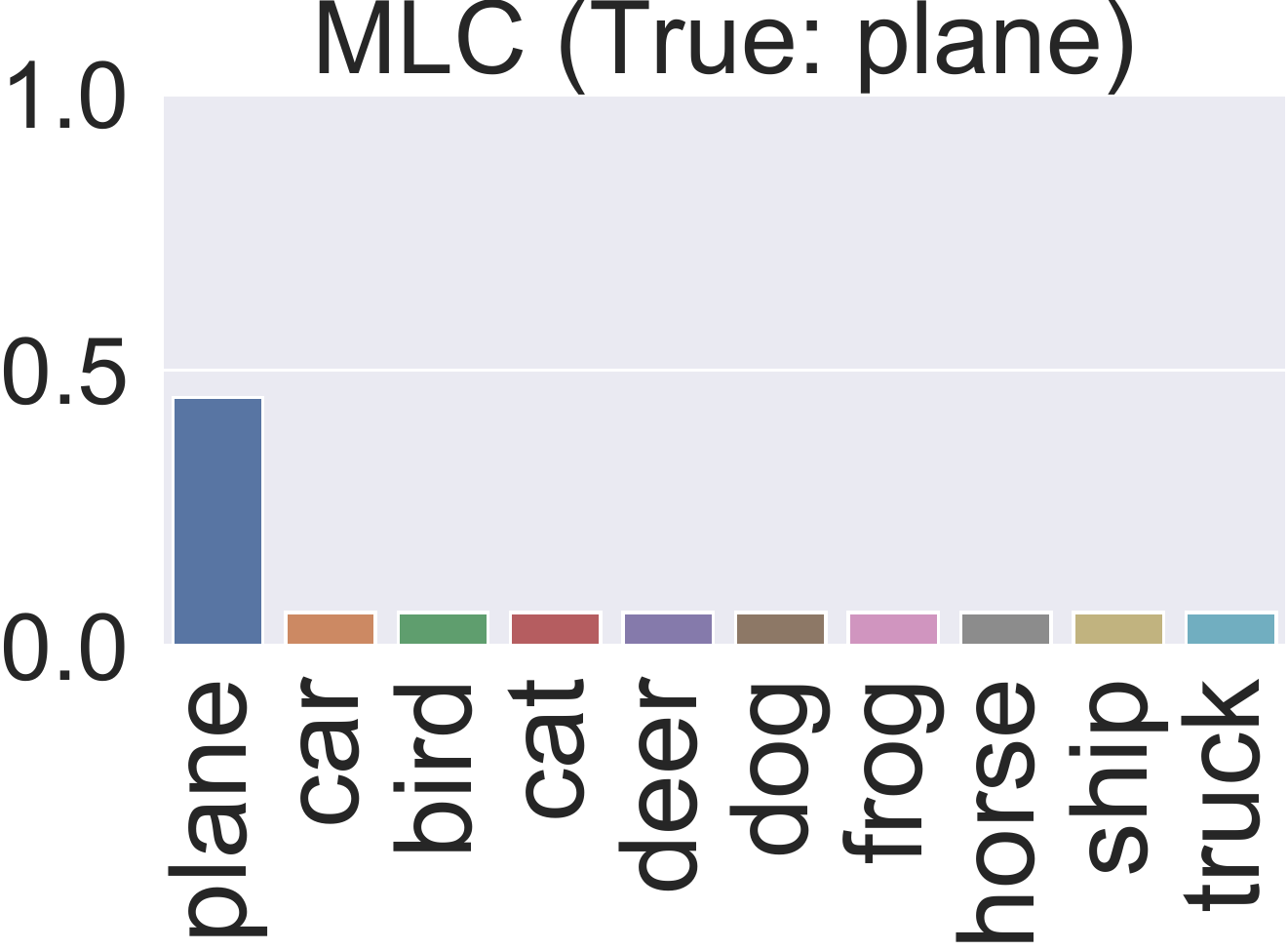}
  \caption{Illustration of label reweighting vs label correction. The first image is a \texttt{truck} and the given noisy label is \texttt{truck}, where MW-Net (A label reweighting method) and MLC successfully recover it. However, for the second image whose true label is \texttt{dog} and the given noisy label is \texttt{automobile}, MW-Net is unable to down-weight the incorrect label while MLC correctly adjusts its label by predicting the maximum weight on the class \texttt{dog} than all other classes. So is the case for the \texttt{airplane}  with noisy label \texttt{frog}.}
  \label{fig:demo}
    \vspace{-0.2in}
  \end{figure}

Training deep networks with noisy labels is challenging since they are prone to fitting and memorizing the noise~\cite{zhang2016understanding} given their high model capacity. As such, multiple lines of work have
been proposed recently to effectively combine clean (or gold) labeled
data with noisy (or weak) supervision data for more effective learning. One
line of work focused on selecting samples from the noisy data that are
likely to be correct using co-teaching or curriculum learning
~\cite{jiang2017mentornet,han2018co}. Another line of work tries to re-weight the weak instances for selective training ~\cite{ren2018learning,shu2019meta}, instead of either including or excluding them. Some of these approaches use a meta-learning framework to assign importance scores to each sample in the noisy training set such that the ones with higher weights can contribute more to the main model training ~\cite{ren2018learning,shu2019meta}.

One of the limitations of label re-weighting is that it is limited to up or down weighting the contribution of an instance in the learning process. An alternative approach relies on the idea of label correction. It aims to correct the noisy labels based on certain assumptions about the weak label generation process. In a sense, label correction aims to go beyond selecting or assigning high weights to useful examples to also altering the assigned labels of incorrectly labeled examples. 
However, previous methods of label correction rely on assumptions about the weak label generation process
and thus often involves two independent steps:  (1) estimating a label corruption
matrix~\cite{hendrycks2018using}, (2) training a model on the noisy data leveraging the corruption matrix. Estimating the corruption matrix often involves assumptions about the noise generation process, such as assuming that the noisy label is only dependent on the true label and is independent of the data itself~\cite{hendrycks2018using}.

In this paper, we adopt label correction to address the problem of
learning with noisy labels, \textit{from a meta-learning
  perspective}. We term our method meta label correction
(MLC). Specifically, we view the label correction procedure as a
meta-process, which objective is to provide corrected
labels for the examples with noisy labels. Meanwhile, the main
predictive model is trained with such corrected
labels generated by the meta-model. Both the meta-model and the main
model are learned concurrently via a bi-level optimization procedure. This allows the model to maximize the performance on the
clean data set (i.e., the clean labels serve as a validation set
w.r.t. the noisy set) by updating the label correction process in a
differentiable manner. MLC extends work on re-weighting and correction
leveraging the advantages of both approaches. In contrast to
meta-learning based instance re-weighting, which only considers up or
down weighting the given noisy label,
MLC provides a more refined way of leveraging noisy
labels by exploring all possible classes in the label space. In
contrast to previous label correction methods, MLC doesn't make
assumptions about the underlying label noises and concurrently learns
a correction model with the main model. Figure \ref{fig:demo} shows
examples where label re-weighting could at best down-weight noisy
samples, reducing their impact on the learning process. On the other
hand, MLC can successfully correct the noisy labels to the true ones.

Meta learning has been successfully used for many applications
including hyper-parameter tuning~\cite{maclaurin2015gradient},
optimizer learning~\cite{DBLP:conf/iclr/RaviL17}, model
selection~\cite{pedregosa2016hyperparameter}, adaptation to new
tasks~\cite{finn2017model} and neural architecture
search~\cite{liu2018darts}. This work leverages meta-learning for
label correction to learn from noisy labels and makes the following
contributions:
\begin{itemize}
\item We pose the problem of learning from weak (noisy) supervision as a meta label correction where a 
  correction network is trained as a meta process to provide reliable labels for the main models to learn;
 \item We compare and contrast re-weighting and correction as two strategies for handling noisy labeled data;
\item We conduct experiments on a combination of 3 image recognition  and 4 large-scale text classification tasks with varying noise levels and types, including real-world noisy labels. We show that  the proposed method outperform previous best methods on label
  correction and re-weighting, demonstrating the power of the proposed method.
\end{itemize}

\section{Related Work}
\label{sec:relatedwork}


Labeled data largely determines whether a machine learning system can
perform well on a task or not, as noisy label or corrupted labels
could cause dramatic performance drop~\cite{nettleton2010study}. The
problem gets even worse when an adversarial rival intentionally
injects noises into the labels~\cite{reed2014training}. Thus,
understanding, modeling, correcting, and learning with noisy labels
has been of interest at large in the research
communities~\cite{natarajan2013learning, frenay2013classification}.
Several approaches ~\cite{mnih2012learning, patrini2017making,sukhbaatar2014training, larsen1998design} have
attempted to address the weak labels by modifying the model's
architecture or by implementing a loss
correction. ~\cite{sukhbaatar2014training} introduced a stochastic
variant to estimate label corruption, however the method has to have
access to the true labels, rendering it inapplicable when no true
labels are present. A forward loss correction adds a linear layer to
the end of the model and the loss is adjusted accordingly to
incorporate learning about the label noise. ~\cite{patrini2017making}
also make use of the forward loss correction mechanism, and propose an
estimate of the label corruption estimation matrix which relies on
strong assumptions, and does not make use of clean labels that might
be available for a portion of the data set. Similar idea is also
explored in ~\cite{goldberger2016training}.

In this paper, we limit our attention to the setting where in addition
to a large amount of weakly labeled data, there is also a small set of
clean data available.  Under this setup, two major lines of work have
been proposed to solve learning problem with noisy labels and we
briefly review them here.

\subsection{Learning with Label Correction}

The first line of work aims to correct the weak labels as much as
possible by imposing assumptions of how the noisy labels are generated
from its underlying true labels. Consider the problem of classifying
the data into $k$ categories, label correction involves estimating a
label corruption matrix $C_{k\times k}$ whose entry $C_{ij}$ denotes
the probability of observing noisy label for class $i$ while the
underlying true class label is actually
$j$~\cite{han2018masking,yao2020dual,xia2019anchor}. For example, gold
loss correction~\cite{hendrycks2018using} falls into this category; a
key drawback of this line of work is that the label corruption matrix
is estimated in an adhoc way and also that the estimation process is
separate from the main model process, hence allowing no feedback from
the main model to the estimation process. In addition, the estimated
label corruption matrices are global, thus ignoring data dependent
noises, a setting prevalent in real world label
noises\cite{xia2020part}.


\subsection{Learning to Re-weight Training Instances}

Knowing that not all training examples are equally important and
useful for building a main model given the noise, another line of work
for learning with weak supervision focuses on selecting a subset of
samples from the noisy data that are likely to be correct
~\cite{jiang2017mentornet,han2018co,pmlr-v97-yu19b,fang2020rethinking}. Instead of discarding examples,
an extension of this idea focused on assigning learnable weights to
each example in the training noisy set. The goal is to assign a weight
for each training example, indicating how useful the example is, such
that the main model could use these weights to improve performance on
a separate validation set (the clean set) ~\cite{ren2018learning, shu2019meta}. The
example weights are essentially hyper-parameters for the main model
and can be learned by formulating a bi-level optimization
problem. This framework allows the example weights learning and the
main model to communicate with each other and a better model could be
learned.

Our work follows the \emph{learning to correct} framework by learning
to model and correct the label noise in the noisy examples. Instead of
separately handling the label correction and model learning steps, we
propose a meta-learning approach to co-optimize for the two steps. We
show that our model can outperform state-of-the-art methods for both
learning to correct and learning to re-weight.

\section{Meta Label Correction}
\label{sec:model}

Following~\cite{charikar2017learning, veit2017learning, Li_2017_ICCV, xiao2015learning, ren2018learning}, we assume that the setup of learning with noisy
labels involves two sets of data: a small set of data with
clean/trusted labels and a large set of data with noisy/weak
labels. Typically the clean set is much smaller compared to the noisy
set, due to scarcity of expert labels and high labeling costs. Training
directly on the small clean set often tends to be sub-optimal, as too
little data can easily cause over-fitting. Training directly on the
noisy set (or a combination of the noisy and clean sets) also tends to
be sub-optimal, as large high-capacity models can fit and memorize the
noise~\cite{zhang2016understanding}. Note that unlike some of the work in this area, e.g.,~\cite{veit2017learning}, we do not require having trusted and noisy labels for the same instances.

One advantage of the label correction approach is that it allows us to combine clean labels and \emph{corrected} noisy labels in the learning process. Our proposed approach adopts the label correction methodology
while also co-optimizing the label correction process together with the main model process through a unified meta-learning framework. We
achieve that by training a meta learner (meta model) that tries to correct the noisy labels and a main 
model that tries to build the best predictive model with corrected labels
coming from the meta model, allowing the meta model and main model to reinforce each other.

\begin{figure}[t]
  \centering
  \includegraphics[width=1.02\linewidth]{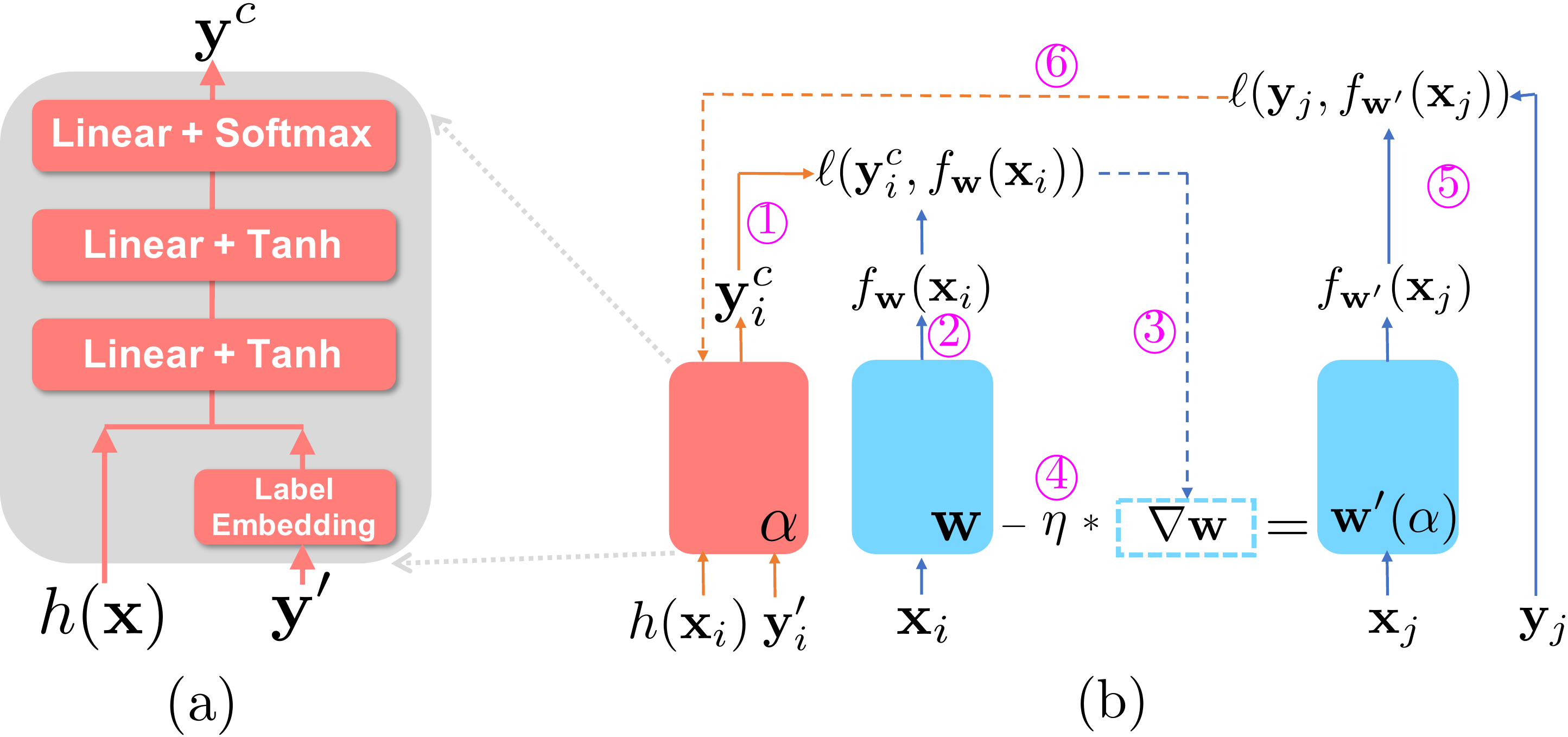}
    \caption{ MLC computation graph. $(\vect x_i, \vect y_i')$ denotes a
      pair of sample with weak label and $(\vect x_j, \vect y_j)$ is a
      pair of sample with clean label. (a) Architecture of the label
      correction network, where $h(\vect x)$ is a feature
      representation of input $\vect x$; (b) Computation flow of
      updating the LCN. In order, operations are:
      \textcolor{black}{\textcircled{1}} Feed the weak instance to the
      LCN and get its corrected label,
      \textcolor{black}{\textcircled{2}} Feed the data instance to the
      current classifier and compute the logits for prediction,
      \textcolor{black}{\textcircled{3}} Compute the loss with the
      logits and corrected label, and compute the gradient of the loss
      w.r.t. the parameter of the classifier. Note that the
      gradient will be a function of the parameters of the LCN.
      \textcolor{black}{\textcircled{4}} Update the classifier
      parameter while keeping the computation graph for its gradient,
      \textcolor{black}{\textcircled{5}} Feed a pair of clean instance
      to the new model and compute its loss,
      \textcolor{black}{\textcircled{6}} Compute the gradient of the
      loss w.r.t the parameter of LCN and update the LCN.}
    \label{fig:model}
\end{figure}


\subsection{A Meta-learning Method for Label Correction}
We describe the framework in detail as follows. Given a set of clean
data examples $D=\{\vect x,\vect y\}^m$ and a set of weak (noisy) data
examples $D'=\{\vect x, \vect y'\}^M$ with $m$ much smaller than
$M$. To best exploit the information carried by the weak labels, we
propose to construct a label correction network (LCN), serving as a
\textit{meta model}, which takes a pair of noisy data example and its
weak label as input and attempts to produce a corrected label for this
data example. The LCN is parameterized as a function with parameters
$\vect \alpha$, $y_c=g_{\vect \alpha}(h(\vect x), y')$ to correct the
weak label $y'$ of example feature $h(\vect x)$ to a more accurate one. (Note
that $y_c$ is a soft label, i.e., a multinomial distribution for all possible classes and the subscription in $y_c$ emphasizes that it's generating a
corrected label). Meanwhile, the \textit{main model} $f$, that we aim
to train and use for prediction after training, is instantiated as
another function with parameters $\vect w$, $y=f_{\vect w}(\vect x)$.

Without linking the two models, there's no way to enforce that: 1) the
corrected label from LCN for an example from the meta model $g$ is
indeed a meaningful one, let alone a corrected one, since directly
training the LCN is not possible without clean labels for the noisy
examples ; 2) The main model $f$ ends up fitting onto the correct true
labels, if the labels provided by the LCN do not align with the
unknown true labels. Fortunately, the two models can be linked
together via a bi-level optimization framework, motivated by the
intuition that \textit{if the labels generated by the LCN are of high
  quality, then a classifier trained with such corrected labels as
  supervision should achieve low loss on a separate set of clean
  examples}. Formally, this can be formulated as the following bi-level optimization problem:
\begin{align}
  \label{eq:opt}
  &\min_{\vect\alpha}\,\mathbb{E}_{(\vect x, y)\in D}\,\ell\left(y,f_{\vect w^*_{\vect \alpha}}(\vect x)\right)\\
  &
  \,\,\,\,\mbox{s.t. }\,\,\vect w^*_{\vect \alpha}=\arg\min_{\vect w}\mathbb{E}_{(\vect x, y')\in D'}\,\ell\left(g_{\vect \alpha}(h(\vect x), y'), f_{\vect w}(\vect x)\right)\nonumber
\end{align}
where $\ell(\cdot)$ is the loss function for classification, i.e.,
cross-entropy\footnote{Note that cross-entropy loss also works with
  soft labels in the lower-level optimization of Problem
  (\ref{eq:opt})}. We term this framework as Meta Label Correction
(MLC); Figure \ref{fig:model} provides an overview of the
framework. Note that to facilitate a light-weight design of the LCN,
we take $h(\vect x)$ to be the feature representations from the main
classifier, e.g., representations from the last layer, with
\textbf{stop-gradient operators} before feeding to LCN to prevent
gradient flowing the LCN back to the main model.

In this bi-level optimization, the LCN parameters $\vect \alpha$ are
the upper parameters (or meta parameters) while the main model
parameters $\vect w$ are the lower parameters (or main
parameters). Like many other work involving bi-level optimizations,
exact solutions to Problem (\ref{eq:opt}) requires solving for the
optimal $\vect w^*$ whenever $\vect \alpha$ gets updated. This is
 both analytically infeasible and computationally expensive,
particularly when the main model $f$ is complex, such as
ResNet~\cite{he2016deep} for image recognition and
BERT~\cite{devlin2018bert} for text classification.

\begin{figure}[t]
  \centering 
  \includegraphics[height=0.31\linewidth]{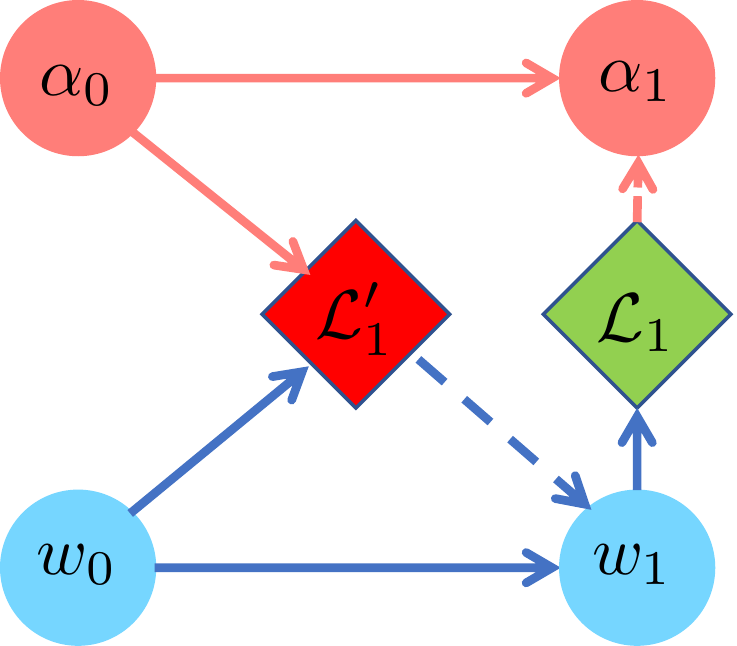}
  \hfill
  \includegraphics[height=0.31\linewidth]{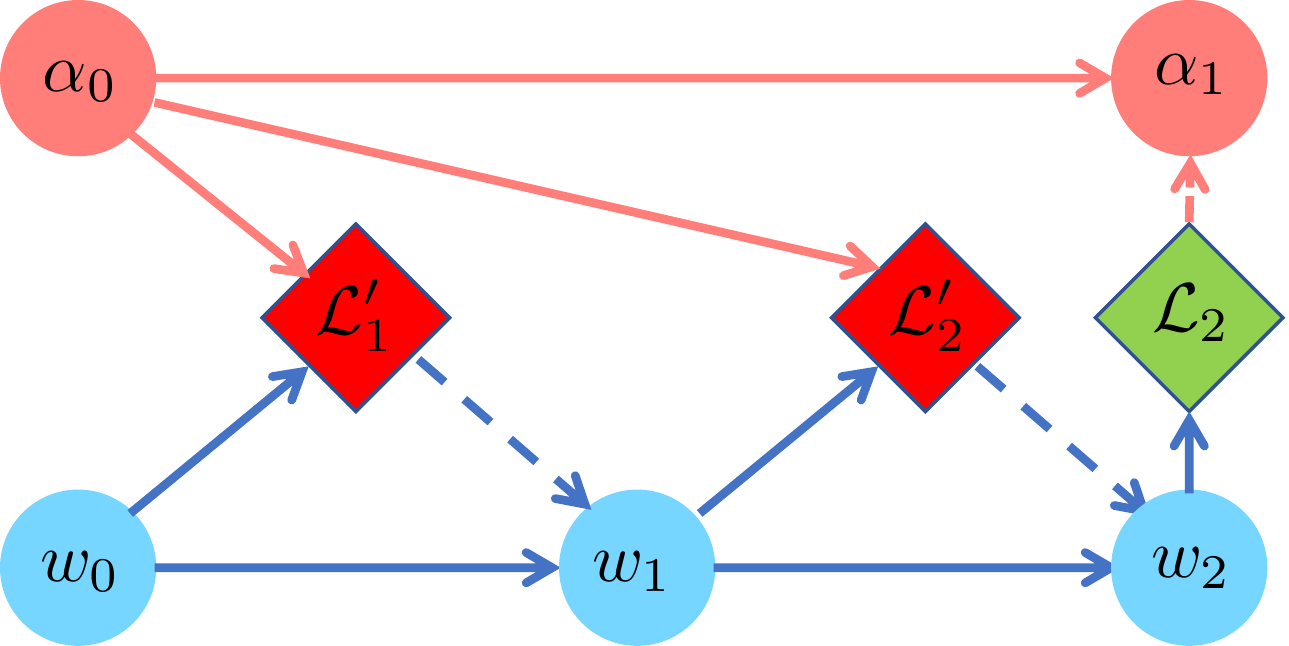}
  \hfill
  \includegraphics[height=0.32\linewidth]{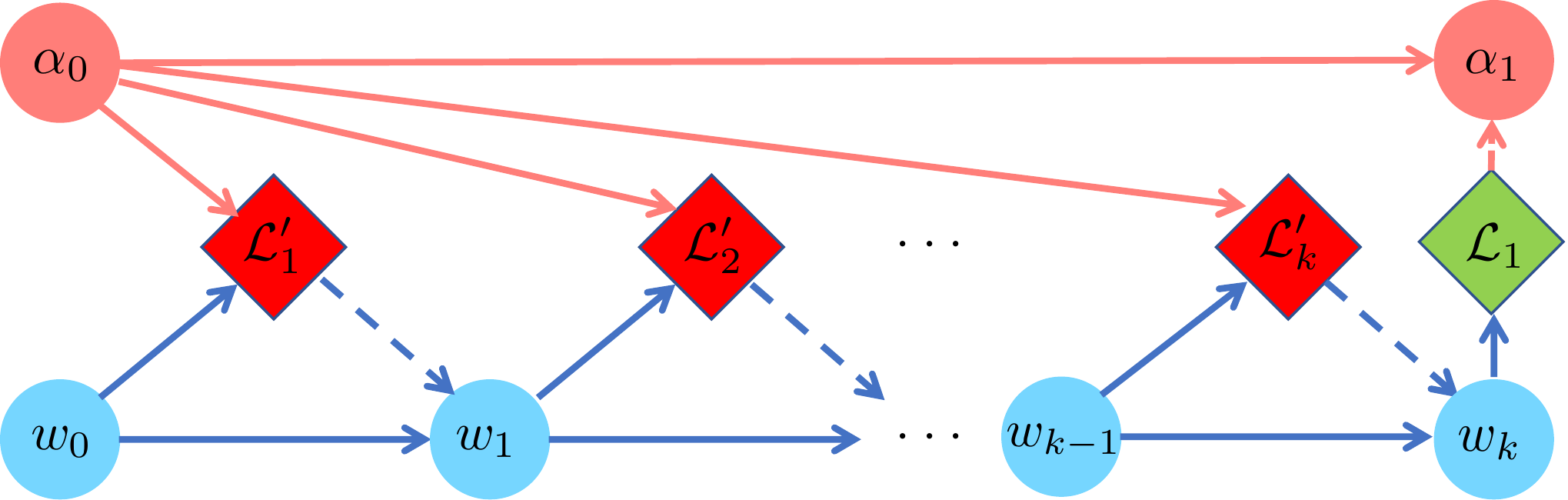}
  \caption{Different bi-level optimization learning strategies.  (a) For any step, the optimal $\vect
    w^*_{\vect\alpha}$ is approximated with one-step look-ahead
    optimizer update, and used to compute the
    evaluation loss $\mathcal{L}_1$, which is then used to update $\vect\alpha$.  (b) $\vect\alpha$ are updated once every two updates from $\vect w$, i.e., a two-step look-ahead update is
    used to approximate $\vect w^*_{\vect
      \alpha}$. (c) Extending to $k$-step look-ahead approximation of
    $\vect w^*_{\vect \alpha}$ for updating $\vect \alpha$.
    We find
    using a $k$ in the range of $1\sim 10$ works well empirically.}
  \label{fig:sgdT}
\end{figure}

\textbf{Gradient-based optimization for bi-level
  optimization}. Outside of label correction research, various other
studies, including differentiable architecture
search~\cite{liu2018darts}, few-shot meta
learning~\cite{finn2017model,nichol2018first}, have used similar bi-level formulation
as Problem (\ref{eq:opt}). Instead of solving for the optimal for
$\vect w^*$ for each $\vect \alpha$, one step of SGD update for $\vect
w$ to approximate the optimal main model for a given $\vect \alpha$
has been employed\footnote{For clarity, we derive this with plain SGD,
  however this also holds for most variants of SGD, including SGD with
  momentum, Adam~\cite{kingma2014adam}.}
\begin{align}
\vect w^*_{\vect \alpha}\approx   \vect w'(\vect \alpha)= \vect w-\eta\nabla_{\vect w}\mathcal{L}_{D'}(\vect \alpha, \vect w)
\end{align}
 where $\mathcal{L}_{D'}(\vect \alpha, \vect
w)\triangleq\mathbb{E}_{(\vect x, y')\in D'}\,\ell\left(g_{\vect
  \alpha}(\vect x, y'), f_{\vect w}(\vect x)\right)$ is a shorthand for the
lower-level objective function and $\eta$ is the learning rate for the
main model $f$. Denoting the upper-level objective function (meta loss) as
$\mathcal{L}_{D}(\vect w)\triangleq\vect w$ $\mathbb{E}_{(\vect x,
  y)\in D}\,\ell(y,f_{\vect w}(\vect x))$, the proxy optimization
problem with one-step look ahead SGD now becomes
\begin{align}
  \min_{\vect\alpha}\,\mathcal{L}_{D}(\vect w'(\vect \alpha))&=\mathcal{L}_{D}(\vect w-\eta\nabla_{\vect w}\mathcal{L}_{D'}(\vect \alpha, \vect w))
\end{align}

\subsection{Efficient Meta-gradient with $k$-step SGD of Main Parameters}

Different from DARTS~\cite{liu2018darts}, the meta loss
$\mathcal{L}_{D}(\vect w^*_{\vect \alpha})$ depends only implicitly on the meta parameters $\vect
\alpha$ via the trained model $\vect w^*_{\vect \alpha}$, hence a more
accurate estimate of the optimal solution $\vect w^*_{\vect \alpha}$
for the current LCN parameters $\vect\alpha$ is desired. To this end, we propose
to employ a $k$-step ahead SGD update as the proxy estimate for the
optimal solution. Figure \ref{fig:sgdT} demonstrates the parameter
updating schemes for different $k$. A larger $k$ principally provides
less noisy estimate for the optimal solution $\vect
w^*_{\vect\alpha}$, however it also results in longer dependencies
over the past $k$ iterations, which requires caching $k$ copies of the
model parameters $\vect w$. To address this, we further propose to approximate the
current meta-parameter gradient with information from the previous $k$
step as follows
\begin{align}
  \frac{\partial \vect w'}{\partial \vect\alpha}&=\left(I-\Lambda H_{\vect w,\vect w}\right)\frac{\partial \vect w}{\partial \vect\alpha}-\Lambda H_{\vect\alpha, \vect w}\\
  g_{\vect w'}\frac{\partial \vect w'}{\partial \vect\alpha}&=g_{\vect w'}\left(I-\Lambda H_{\vect w,\vect w}\right)\frac{\partial \vect w}{\partial \vect\alpha}-g_{\vect w'}\Lambda H_{\vect\alpha, \vect w}\\
  \frac{\partial \mathcal{L}_D(\vect w')}{\partial \vect\alpha}&\approx g_{\vect w'}\left(I-\Lambda \right)\frac{g_{\vect w^\top}}{\|g_{\vect w}\|^2}\frac{\partial \mathcal{L}_D(\vect w)}{\partial \vect\alpha}-g_{\vect w'}\Lambda H_{\vect\alpha, \vect w}
 \label{eq:mgradient}
\end{align}
where $\vect w'$ is the model parameter for next step, $g_{\vect w}$
is a short hand for the gradient of the training loss w.r.t $\vect w$,
$\Lambda$ is a diagonal matrix representing the current learning rates
for all parameters in $\vect w$, and $H_{\vect \alpha, \vect w}$ is a
short hand for $\frac{\partial^2}{\partial\vect\alpha\partial\vect
  w}\mathcal{L}_{D'}(\vect\alpha, \vect w)$. $H_{\vect w,\vect w}$ is
estimated with identity to ease computation and the second term can be
computed as
\begin{align}
  g_{\vect w'}\Lambda H_{\vect\alpha, \vect w} =& \nabla_{\vect \alpha, \vect  w}^2\mathcal{L}_{D'}(\vect \alpha, \vect w)\Lambda\nabla_{\vect w'}\mathcal{L}_{D}(\vect w')\nonumber\\
  =&\nabla_{\vect \alpha}\Big(\nabla_{\vect  w}^\top\mathcal{L}_{D'}(\vect \alpha, \vect w)\Lambda\nabla_{\vect w'}\mathcal{L}_{D}(\vect w')\Big)
\end{align}

Algorithm \ref{alg} outlines an iterative procedure to solve the above
proxy problem with $k$-step look ahead SGD for the main model.

\begin{algorithm}[t]
 \caption{MLC - Meta Label Correction}
 \While{not converged}{
   Update meta parameters $\vect \alpha$ by descending Eq. (\ref{eq:mgradient})\\
   Update model parameters $\vect w$ by descending $\nabla_{\vect w}\mathcal{L}_{D'}(\vect \alpha, \vect w)$
 }
 \label{alg}
\end{algorithm}

\subsection{Training with Soft Labels from LCN}

Not only does the LCN explicitly model the dependency of the corrected
label on both the data example and its noisy label, but also it
ensures that the output from the LCN is a valid categorical
distribution over all possible classes. Soft labels are crucial in MLC
as they make gradient propagation back to the meta model from the main
model possible. However, when the main model takes these examples with
soft corrected labels, it brings difficulty to training due to the
additional uncertainly in the corrected labels. This can be alleviated
by the following strategy. In training for each batch of clean data,
we split it into two parts, with one serving as the clean evaluation
set and add the other to the training process for $f$, as a small
portion of the clean set will provide clean guidance for training, to
ease model training. This has been shown to be effective in similar
settings~\cite{ranzato2015sequence, pham2020meta}. 

\subsection{Remark: Label Correction vs Label Reweighting}

\begin{figure}[t]
  \centering
  \begin{subfigure}{0.495\linewidth}
    \centering
    \includegraphics[height=0.7\linewidth]{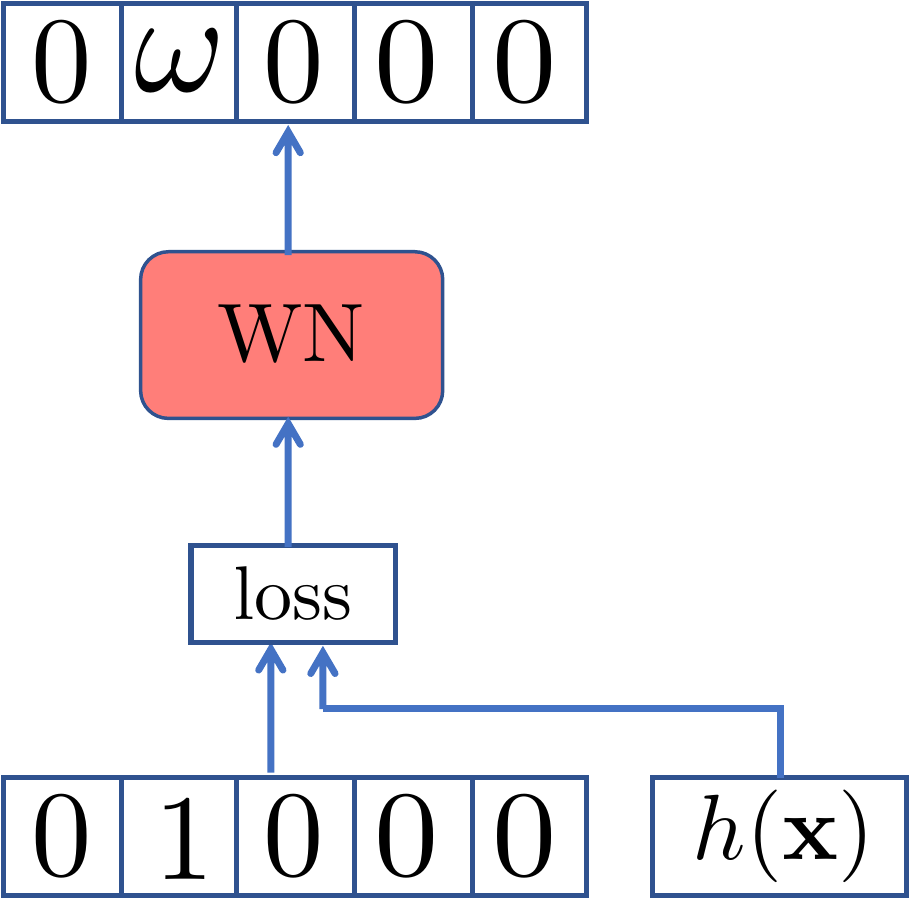}
    \caption{Weighting-Network (WN) in Meta-WN~\cite{shu2019meta}}
  \end{subfigure}
  \hfill
  \begin{subfigure}{0.495\linewidth}
    \centering
    \includegraphics[height=0.7\linewidth]{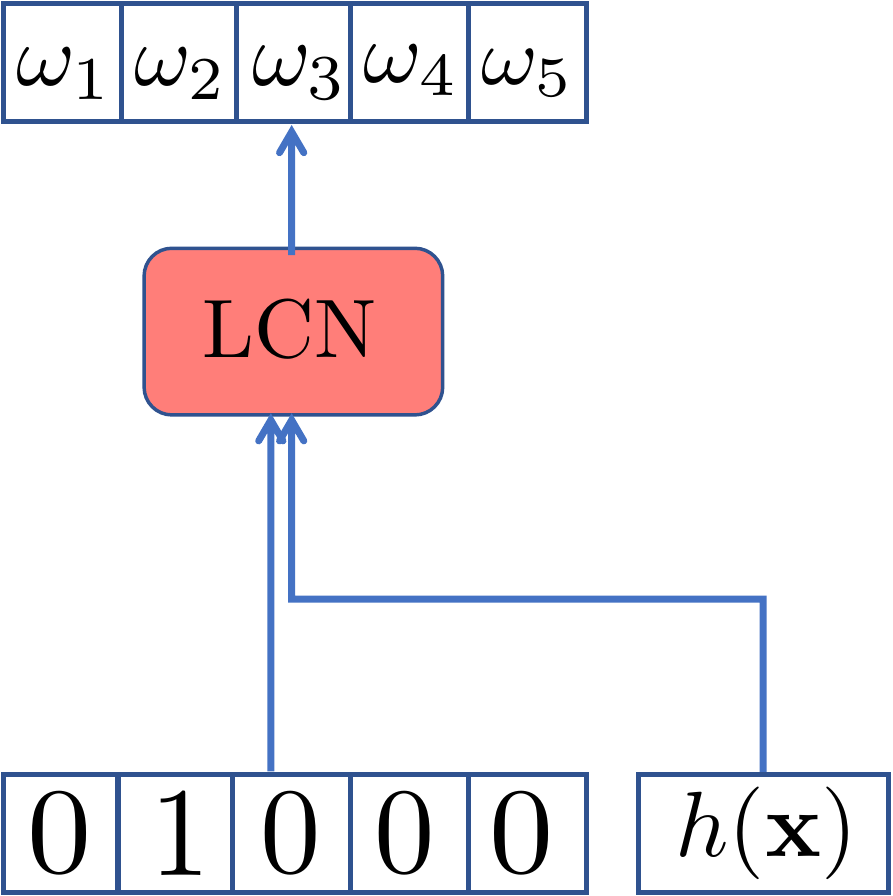}
    \caption{Label Correction Network (LCN) in MLC (Ours)}
  \end{subfigure}
  \caption{Different treatments of noisy labels from Meta-WN and
    MLC. Total number of classes assumed to be 5 for illustration
    purpose. (a) Meta-WN; (b) MLC.}
  \label{meta_ops}
  \vspace{-0.15in}
\end{figure}

To address noisy labels, Meta-WN~\cite{shu2019meta} leverages the
Weight-Network (WN) as the meta-module to \textbf{reweight} the given
noisy label, while MLC aims to provide a more refined treatment, i.e.,
to \textbf{correct} the noisy label. More explicitly, Figure
\ref{meta_ops} demonstrates the difference of the underlying
operations between Meta-WN and MLC. To highlight
\begin{itemize}
  \item For an input, Meta-WN tries to learn a weight for the given
    noisy class only, whiling ignoring all other possible classes
    (demonstrated by the single non-negative weight $\omega$ for Class
    2 while $0$ for all other classes in Figure \ref{meta_ops}(a)),
    while since MLC tries to correct the given weak label, essentially
    it considers all possible classes (demonstrated by the full
    non-negative vector resulting from the final softmax layer of the LCN,
    i.e., $(\omega_1, \omega_1, \omega_3, \omega_4, \omega_5)$ for all
    classes with $\omega_1+ \omega_1+ \omega_3+ \omega_4+ \omega_5=1$,
    essentially weighting all possible classes)
  \item Another key difference between MLC and Meta-WN relies on the
    information bottleneck to their corresponding meta-modules. For
    Meta-WN doesn't directly take an pair of data and noisy label as
    input, but rather relies on the \textit{scalar} loss of the
    classifier that particular input incurs as input. In other words,
    the meta-module lacks the ability to differentiate the different
    input pairs if the loss values for them are similar, effectively
    limiting the modeling capacity of the meta-module. While the LCN
    in MLC directly takes the data feature $h(\vect x)$ and its weak
    label as input, allowing a more flexible treatment and enabling
    the LCN to identify different information brought by different
    input pairs.
  \end{itemize}

\section{Experiments}
\label{sec:exp}

To test the performance of MLC, we conduct experiments on a
combination of three image recognition and four text classification
tasks, and compare with previous state-of-the-art approaches for
learning with noisy labels under different types of label noises.

\subsection{Datasets and Setup}
\textbf{Datasets}. We evaluate our method on 3 image recognition datasets, CIFAR-10, CIFAR-100~\cite{krizhevsky2009learning} and
Clothing1M~\cite{xiao2015learning} and 4 large-scale multi-class text classification benchmark
datasets, that are widely used by text classification
research~\cite{zhang2015character, xie2019unsupervised,
  dai2019transformer, yang2016hierarchical, conneau2016very}, AG news, Amazon reviews, Yelp reviews and Yahoo
answers. Information about all datasets is summarized in Table
\ref{tab:data}.

\begin{table*}[h]
  \small
  \centering
  \begin{tabular}{l|rr|r|rrrr}\toprule
    \tb{Dataset} & \tb{CIFAR-10} & \tb{CIFAR-100} & \tb{Clothing1M} & \tb{AG} & \tb{Amazon-5} & \tb{Yelp-5} & \tb{Yahoo}\\
    \midrule
    \# classes & 10 & 100 & 14 & 4 & 5 & 5 & 10\\
    Train & 50K & 50K & 1.05M & 120K & 3M & 650K & 1.4M \\
    Test  & 10K & 10K & 10K & 7.6K & 650K & 50K & 60K \\\midrule
    Clean  & 1000 & 1000 & 50K & 400 & 500 & 500 & 1000 \\    
    Noisy  & 49K & 49K & 1M & 119.6K & $\sim$ 3M & 649.5K & $\sim$ 1.4M \\    \midrule
    Classifier & \multicolumn{2}{c|}{ResNet 32} & ResNet 50 & \multicolumn{4}{c}{Pre-trained BERT-base}\\
    \bottomrule
  \end{tabular}
  \vspace{-0.1in}
      \caption{Dataset statistics and classifier architectures
      used. Note the clean set is significantly smaller than the noisy
      label set.}
  \label{tab:data}        
\end{table*}
\begin{table*}[h]
  \small
  \centering
  \begin{tabular}{l|cc|cccc}\toprule
   \textbf{Datasets} & \textbf{CIFAR-10} & \textbf{CIFAR-100} &\textbf{AG} & \textbf{Yelp-5}& \textbf{Amazon-5} & \textbf{Yahoo}\\
   ($\#$ clean labels) & ($10\times 100$) & ($100\times 10$)& ($4\times 100$)       &  ($5\times 100$)           &  ($5\times 100$)        &  ($10\times 100$)     \\\midrule
   MW-Net~\cite{shu2019meta} & 65.12 & 39.96 & 75.91&  51.27& 49.49 & 60.18     \\
   GLC~\cite{hendrycks2018using} & 86.62 & 50.50 & 83.88 & 60.12 & 60.31 & 68.03\\
   MLC (Ours)  & \bf{86.81} & \bf{53.68} &\bf{85.27}&  \bf{62.61}&  \bf{61.21}&  \bf{73.72}      \\ 
  \bottomrule
  \end{tabular}
    \vspace{-0.1in}
\caption{Mean accuracies on all data sets. Each cell represents the
  average runs over two noise types and 10 noise levels. A $k=5$
  (5-step ahead SGD) is used for all experiment. (Each configuration
  is run for 5 times and the mean is reported)}
\label{tab:summary}
\end{table*}
\begin{table*}[h!]
  \small
  \centering
\begin{tabular}{lcccccc}\toprule
  \textbf{Method} & Forward & Joint Learning  & MLNT & MW-Net & GLC &MLC \\
   & \cite{patrini2017making} & \cite{tanaka2018joint}  & \cite{li2019learning} & \cite{shu2019meta} & \cite{hendrycks2018using} & (Ours) \\
  \midrule
  \textbf{Accuracy} & 69.84 & 72.23 & 73.47 & 73.72 &73.69 &\bf{75.78}\\
\bottomrule
\end{tabular}
\vspace{-0.1in}
\caption{Test set accuracies on Clothing1M with real-world noisy labels ($k=5$)}
\label{tab:clothing1m}
\vspace{-0.1in}
\end{table*}

\textbf{Noisy label sources}
Following related work~\cite{hendrycks2018using,ren2018learning,shu2019meta}, for each dataset, we sample a
portion of the entire training set as the clean set (except for
Clothing1M). To ensure a fair and consistent evaluation, we use only
1000 images as clean set for both CIFAR-10 and CIFAR-100, and only 100
instances per class for the four large scale text classification data
sets. The noisy sets are generated by corrupting the
labels of all the remaining data points based on the following two
setting:

\textbf{Uniform label noise} (\textsf{UNIF}). For a dataset with $C$
classes, a clean example with true label $y$ is randomly corrupted to all
possible classes $y'$ with  probability $\frac{\rho}{C}$ and stays in its original
label with probability $1-\rho$. (Note the corrupted label might also
happen to be the original label, hence the label has probability of
$1-\rho+\frac{\rho}{C}$ to stay uncorrupted.)

\textbf{Flipped label noise} (\textsf{FLIP}). For a dataset with $C$
classes, a clean example with true label $y$ is randomly flipped to
\textit{one of the rest $C-1$ classes} with probability $\rho$ and
stays in its original label with probability $1-\rho$.

We vary $\rho$ in the range of $[0,1]$ to simulate different noise
levels for both types. We emphasize that both simulated noise types
make the assumption that given the true label the noisy label doesn't
depend on the data itself. Hence we also evaluate all the methods on
another source of noisy labels:

\textbf{Real-world noisy labels}. Clothing1M~\cite{xiao2015learning} is a dataset where noisy labels for images are devised by leveraging user tags as proxy annotations.  As Clothing1M is the only dataset that comes with real-world noisy labels, we use its original split of clean and noisy sets.

Finally, we also note that regardless of noise types, none of the methods tested in this paper is aware of the label corruption probability $\rho$ nor do they have knowledge about which data sample in the noisy set is actually corrupted.

\subsection{Baseline Methods and Model Architectures}

We focus our evaluation of MLC against state-of-the-art
methods for learning with weak supervision from two different themes,
i.e., ~\cite{hendrycks2018using} for label correction (denoted by GLC
hereafter) and instance re-weighting with meta learning
~\cite{shu2019meta} (denoted by MW-Net hereafter). Note that GLC and MW-Net were shown to consistently outperform other methods such as training on clean data only, cleaning on weak data only, combining clean and weak data, as well as more sophisticated  models for combining the clean and weak labels such as distillation \cite{Li_2017_ICCV} and forward loss correction~\cite{sukhbaatar2014training}. Additionally,
MW-Net was shown to outperform a slightly different variant for
instance re-weighting with meta-parameters~\cite{ren2018learning}. As
such, we do not show results from these methods.

For fair and consistent comparisons, we use the same classifier
architectures for all methods, i.e., ResNet 32 for CIFAR-10 and
CIFAR-100, ResNet 50 pretrained from ImageNet for Clothing1M, and
pre-trained BERT-base for the four large-scale text data sets. We
implement all models and experiments in PyTorch. All models are
trained with the same number of epochs for the same
dataset.\footnote{Code for MLC is available at \url{https://aka.ms/MLC_}}

\textbf{LCN architecture}. We use the same LCN architecture for MLC
across all settings as follows (Figure \ref{fig:model}(a)):
\begin{itemize}
    \item[] An embedding layer of size $(C, 128)$ to embed the input noisy labels,
    followed by a three-layer feed-forward network with dimensions of \texttt{(128+xdim, hdim), (hdim, hdim), (hdim,C)} respectively. \texttt{tanh} is used as the nonlinear activation function in-between them and lastly a Softmax layer to output a categorical distribution as the corrected labels
\end{itemize}
where $C$ is the number of classes, \texttt{xdim} is the feature dimension of input $\vect x$ from
the last layer from the main classifier, i.e., 64 from ResNet32 for CIFAR-10
and CIFAR-100, 2048 from ResNet 50 for Clothing1M and 768 from
BERT-base for text datasets and \texttt{hdim} is the hidden dimension
for the LCN (set to 768 for text datasets and 64 otherwise).

\subsection{Main Results}

\textbf{MLC on image recognition.} We start by comparing all methods
on the standard image recognition datasets. Table \ref{tab:summary}
presents the averaged accuracies across multiple configurations (two
noise types, 10 noise levels) with $k=5$. The table shows that MLC
consistently outperforms other methods over all datasets.  In
addition, on Clothing1M with real noisy labels (Table
\ref{tab:clothing1m}), MLC outperforms all baseline methods and
improves over GLC and MW-Net by over 2 points in accuracy, suggesting
its ability to capture better data-dependent label corruptions via the
meta-learning framework.

\textbf{MLC on text classification.} Table \ref{tab:summary} also
presents the mean accuracies of MLC on 4 large text data sets with
pre-trained BERT-base as its main classifier. Overall, label reweighting (MW-Net) seems insufficient to fully address the text classification problem; label correction approaches demonstrate much higher performances, while MLC achieves the best thanks to its nature of combining of both label
correction and the data-driven meta-learning framework.

\subsection{Analysis and Ablation Studies}

\begin{figure}[t]
   \centering
   \includegraphics[width=0.9\linewidth]{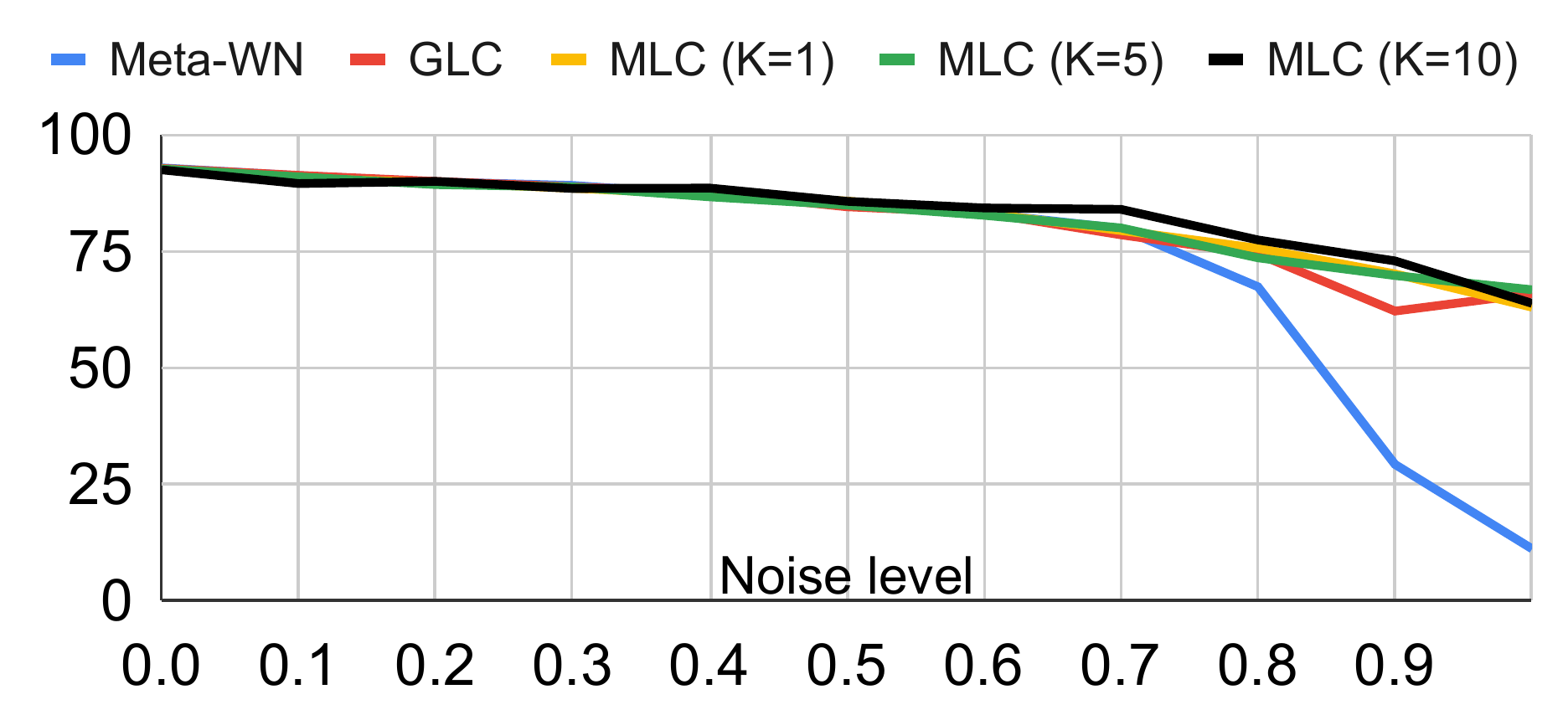}
  \caption{Test accuracy w.r.t noise levels}
  \label{ref:noise_level_unif}
  \end{figure}

\textbf{Effects of noise levels $\rho$ and $k$ for MLC.} Figure \ref{ref:noise_level_unif} presents the results of all
methods under \textsf{UNIF} with noise levels from 0 to 1.0 with a
step size of 0.1. It's clear that, since Meta-WN only attempts to
re-weight the observed weak label, its performance decreases
significantly when the noise level goes up, as the given label turns
more likely to be the wrong label thus re-weighting for this case is
insufficient; while label correction based methods (GLC and MLC) show
to be robust against severe label noises. This is consistent with results reported in ~\cite{hendrycks2018using} where label correction was shown to perform well even in extreme noise level situations. 
Moreover, we observe that MLC is more effective in doing this than previous label correction methods for severe 
noise. In terms of the number of look-ahead steps, $k$, used to compute the meta-gradient, the value of $k$ does not seem to have an impact on MLC's performance when the noise level is low; however when the noise level is high (more than 0.6), a larger $k$ leads to higher test accuracy, validating the strategy of using multiple steps to compute the meta-gradients. Similar trends are also observed on \textsf{FLIP}.

\begin{figure}[t] 
  \centering
  \includegraphics[width=0.49\linewidth]{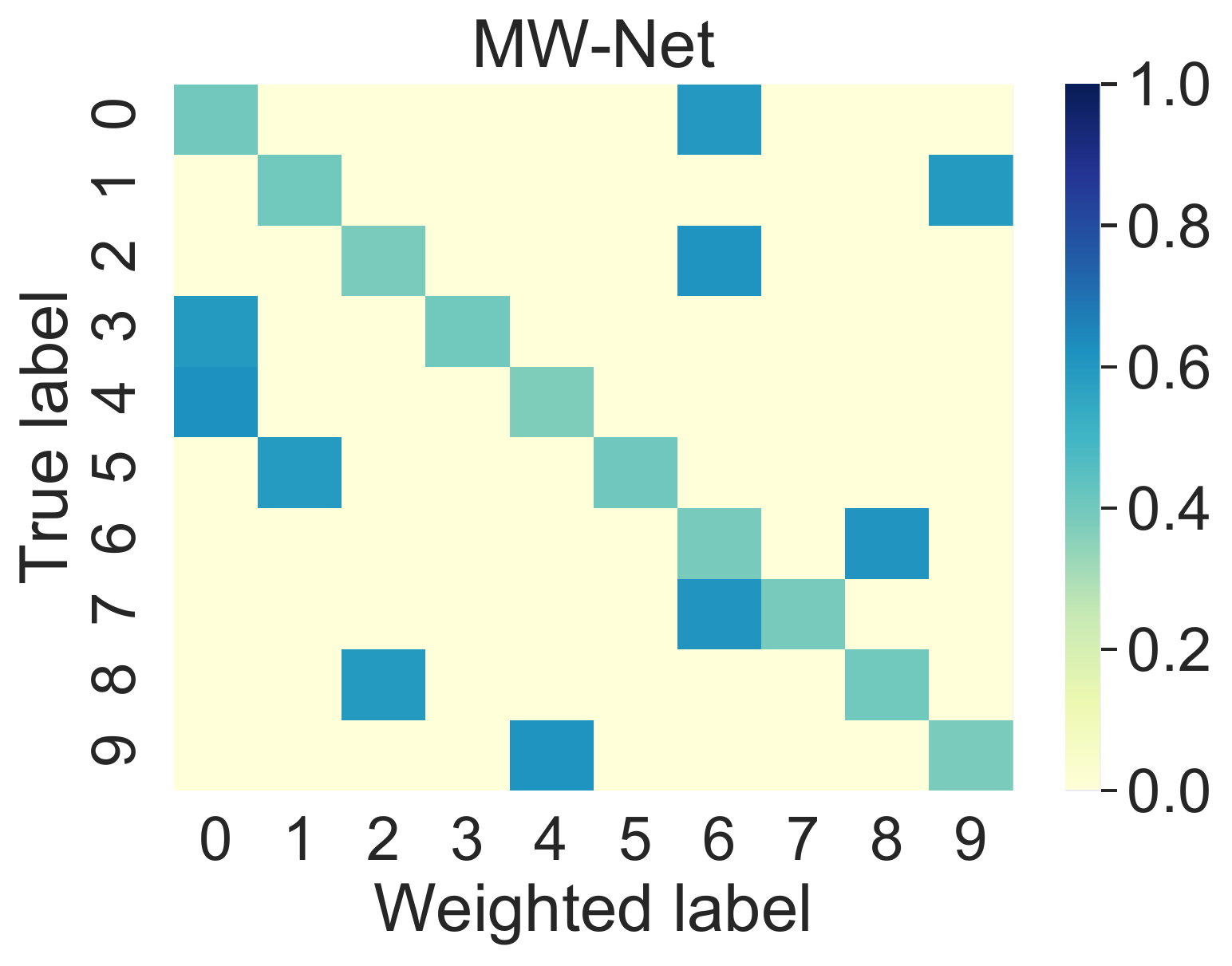}
  \includegraphics[width=0.49\linewidth]{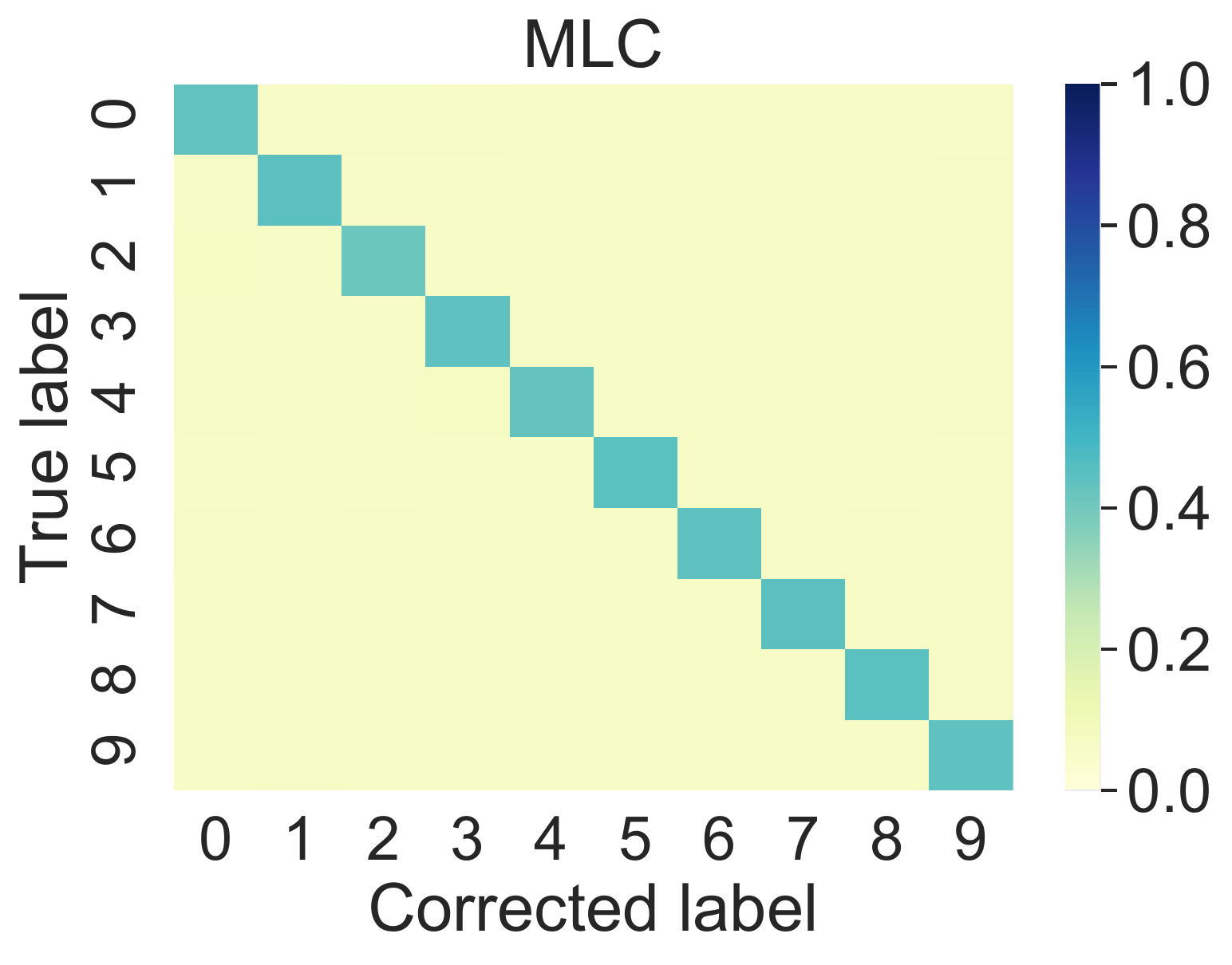}
  \caption{(a) Heatmap of learned weights of MW-Net w.r.t to the true
    labels; note that MW-Net does not alter the noisy labels but only assigns them weight. (b) Heatmap of probability distribution of the corrected
    labels of MLC w.r.t to the true labels; note that MLC can alter the input label (For all examples in the
    test set with noise level $\rho=0.6$.)}
  \label{fig:metanet}
\end{figure}

\begin{figure}[t]
  \centering
\includegraphics[width=0.49\linewidth]{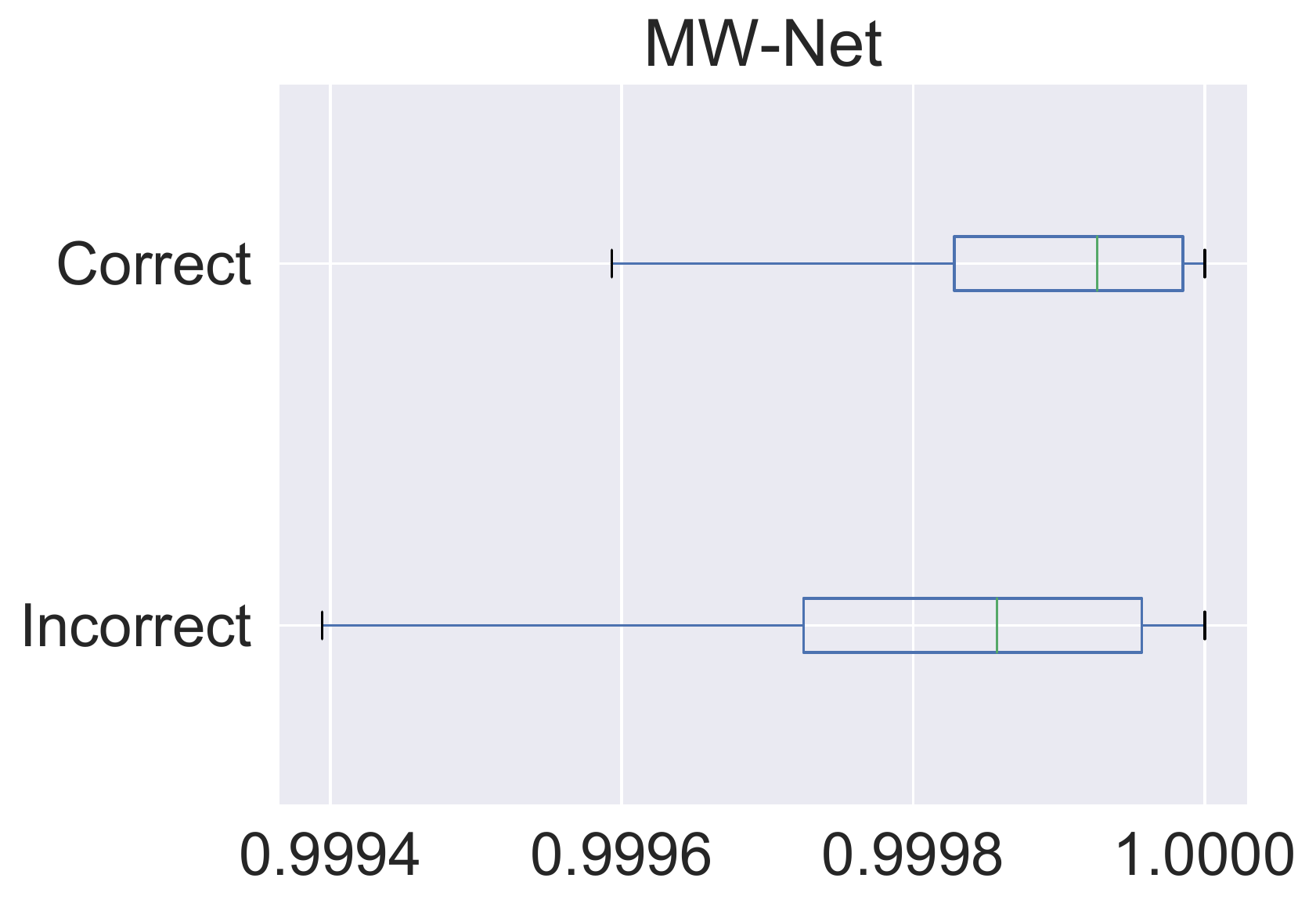}
\includegraphics[width=0.49\linewidth]{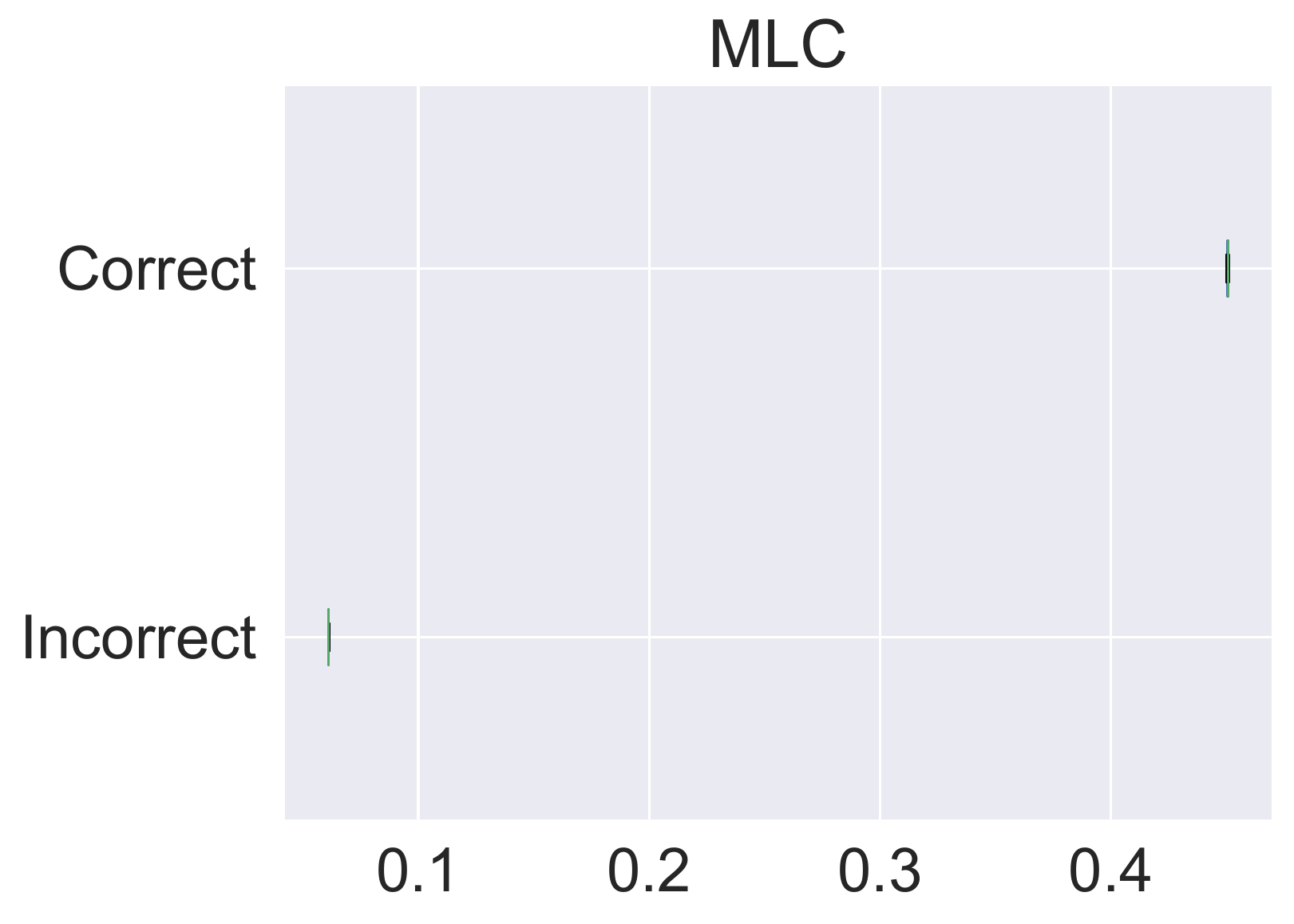}
  \caption{Effect of reweighting vs correction for \textsf{FLIP} with $\rho=0.6$. (Left) Weights assigned by MW-Net w.r.t to the input noisy label; (Right) Correction probabilities learned by LCN to the input noisy label.}
  \label{fig:boxplot}
\end{figure}

\textbf{Meta net evaluation.} We perform additional experiments to understand what the meta model, i.e., the LCN, actually learns
after model convergence. Additionally, we seek to quantify the benefit of correcting noisy labels, v.s. re-weighting instances.  We use the \textsf{FLIP} setting to
generate corrupted labels for instances in the test set in CIFAR-10
and feed them to the meta nets of both MLC and Meta-WN. MLC will produce a probability distribution over all possible classes where Meta-WN will assign a scalar weight to each instance.
Note that, for CIFAR-10, we know which of the noisy label is actually correct and which is not but neither of the models have access to this information. Ideally, Meta-WN will assign higher weights to the correct instances and lower weights to the incorrect ones. Similarly, MLC should keep the label of correct instances as is and alter the labels of the incorrect ones. We see from Figure~\ref{fig:boxplot} that this is actually the case. On average, both model seem to be able to distinguish between the correct and incorrect labels. However, for incorrect labels, Meta-WN can only down-weight the sample reducing the dependence of the training process on it. MLC goes beyond this by also trying to change the label to assign the sample to the correct one, allowing the main model to fully leverage it. We can see from Figure~\ref{fig:metanet} that it does that successfully. On other other hand, MW-Net can only assign a weight to the noisy label. 

\begin{figure}[t]
  \centering
  \begin{subfigure}{0.32\linewidth}
      \centering
    \includegraphics[width=\linewidth]{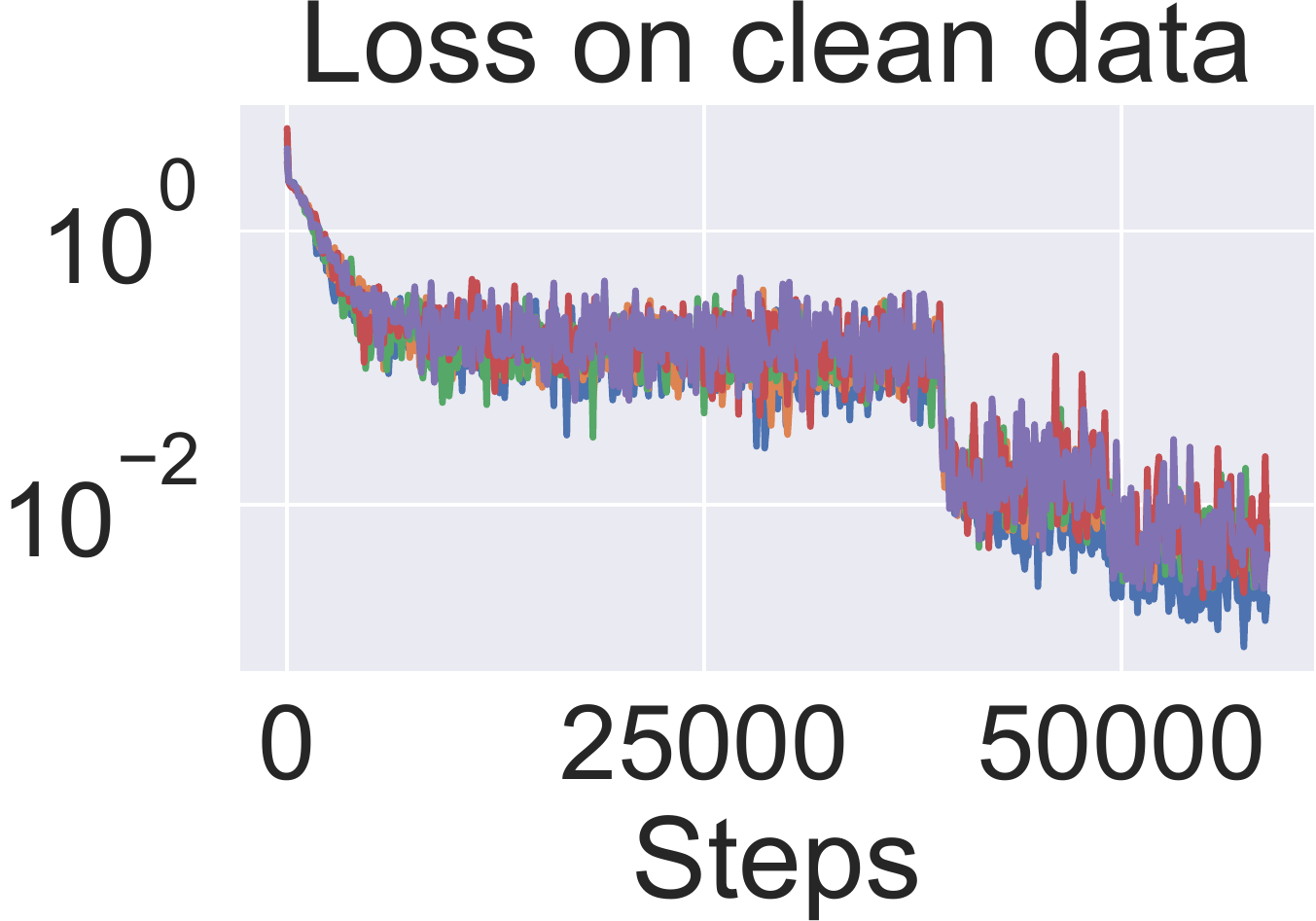}
    \caption{}
  \end{subfigure}
  \begin{subfigure}{0.32\linewidth}
      \centering
      \includegraphics[width=\linewidth]{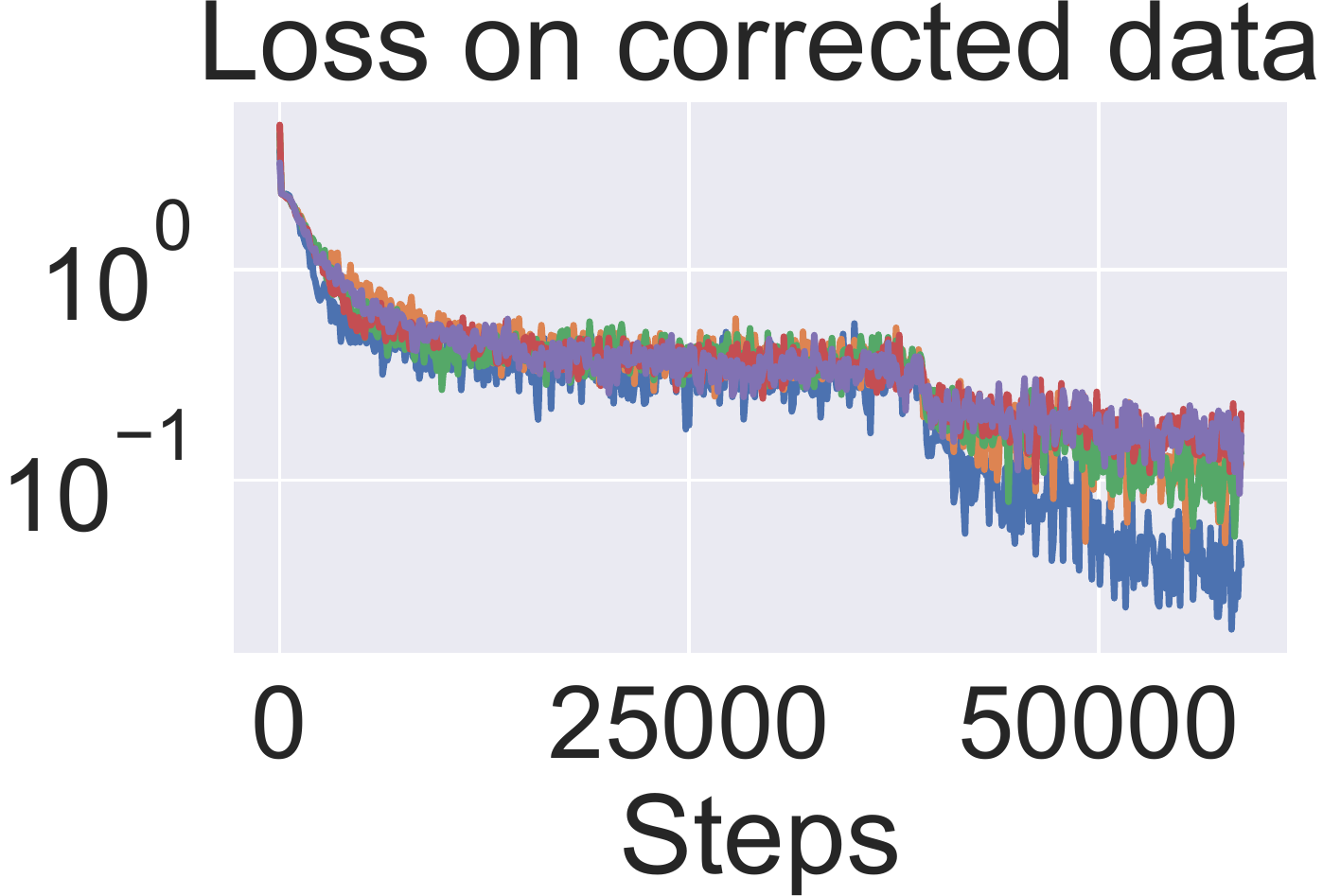}
      \caption{}
    \end{subfigure}
  \begin{subfigure}{0.32\linewidth}
      \centering
      \includegraphics[width=\linewidth]{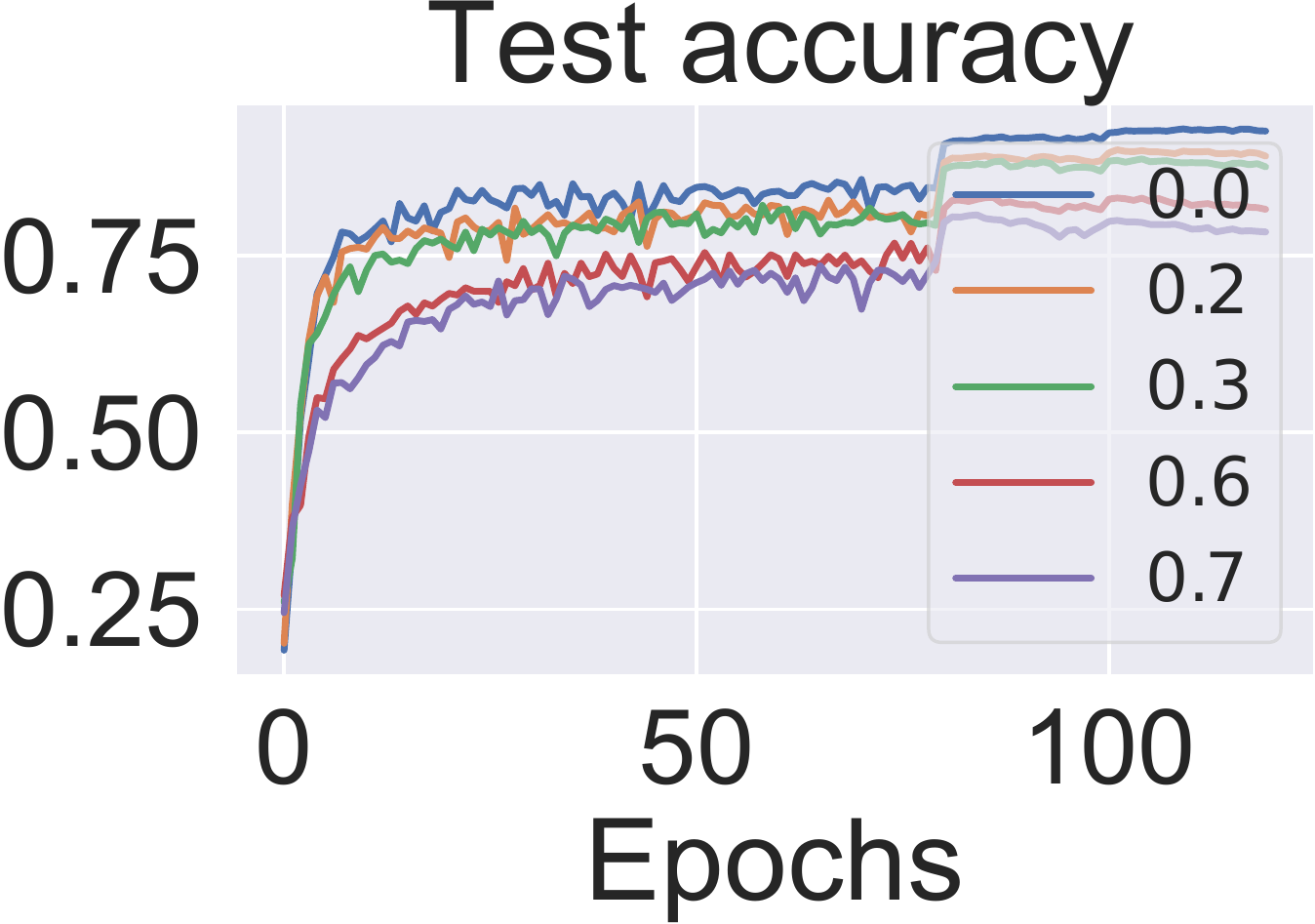}
      \caption{}
  \end{subfigure}
  \caption{(a,b,c) Loss and test set accuracy dynamics w.r.t noise levels (losses are
    in log-scale). The two jump-points in the curves are due to the
    decay of learning rates at 60th and 80th Epoch, or $\sim$ 40K and
    50K steps equivalently.}
  \label{fig:loss}
\end{figure}

\textbf{MLC training dynamics.} Figure \ref{fig:loss}(a,b,c) shows the training progress for one run
on the CIFAR-10 with \textsf{UNIF} under different noise levels. We
monitor a set of different metrics during training, including the loss
function on the noisy data with corrected labels, loss function
on clean data, and the test set accuracy as training progresses. The
figure shows that both losses decrease and test accuracy increases as
the training process progresses. Note that with larger noise levels
(hence more difficult cases), training with MLC gets harder (as seen
by the slightly higher loss on clean data and loss on noisy
data). However, MLC still converges and achieves
good results on test set as shown in Figure \ref{fig:loss}(c). 

\section{Conclusions}
\label{sec:conclusion}

In this paper, we address the problem of learning with noisy labels
from a meta-learning perspective. Specifically, we propose to use a
meta network to correct the noisy labels from the data set, and a main
classifier network is trained to fit the example to a provided label,
i.e., corrected labels for the noisy examples and true labels for the
clean ones. The meta network and main network are jointly optimized in
a bi-level optimization fashion; to address the computation challenge,
we employ a $k$-step ahead SGD update to compute the
meta-gradient. Empirical experiments on three image recognition and
four text classification tasks with various label noise types show the
benefits of label correction over instance re-weighting and
demonstrate the strong performance of MLC over previous methods
leveraging noisy labels.

\bibliography{ref}

\begin{thebibliography}{45}
\providecommand{\natexlab}[1]{#1}
\providecommand{\url}[1]{\texttt{#1}}
\providecommand{\urlprefix}{URL }
\expandafter\ifx\csname urlstyle\endcsname\relax
  \providecommand{\doi}[1]{doi:\discretionary{}{}{}#1}\else
  \providecommand{\doi}{doi:\discretionary{}{}{}\begingroup
  \urlstyle{rm}\Url}\fi

\bibitem[{Charikar, Steinhardt, and Valiant(2017)}]{charikar2017learning}
Charikar, M.; Steinhardt, J.; and Valiant, G. 2017.
\newblock Learning from untrusted data.
\newblock In \emph{Proceedings of the 49th Annual ACM SIGACT Symposium on
  Theory of Computing}, 47--60. ACM.

\bibitem[{Conneau et~al.(2016)Conneau, Schwenk, Barrault, and
  Lecun}]{conneau2016very}
Conneau, A.; Schwenk, H.; Barrault, L.; and Lecun, Y. 2016.
\newblock Very deep convolutional networks for text classification.
\newblock \emph{arXiv preprint arXiv:1606.01781} .

\bibitem[{Dai et~al.(2019)Dai, Yang, Yang, Carbonell, Le, and
  Salakhutdinov}]{dai2019transformer}
Dai, Z.; Yang, Z.; Yang, Y.; Carbonell, J.; Le, Q.~V.; and Salakhutdinov, R.
  2019.
\newblock Transformer-xl: Attentive language models beyond a fixed-length
  context.
\newblock \emph{arXiv preprint arXiv:1901.02860} .

\bibitem[{Devlin et~al.(2018)Devlin, Chang, Lee, and
  Toutanova}]{devlin2018bert}
Devlin, J.; Chang, M.-W.; Lee, K.; and Toutanova, K. 2018.
\newblock Bert: Pre-training of deep bidirectional transformers for language
  understanding.
\newblock \emph{arXiv preprint arXiv:1810.04805} .

\bibitem[{Fang et~al.(2020)Fang, Lu, Niu, and Sugiyama}]{fang2020rethinking}
Fang, T.; Lu, N.; Niu, G.; and Sugiyama, M. 2020.
\newblock Rethinking Importance Weighting for Deep Learning under Distribution
  Shift.
\newblock \emph{Advances in Neural Information Processing Systems} 33.

\bibitem[{Finn, Abbeel, and Levine(2017)}]{finn2017model}
Finn, C.; Abbeel, P.; and Levine, S. 2017.
\newblock Model-agnostic meta-learning for fast adaptation of deep networks.
\newblock In \emph{Proceedings of the 34th International Conference on Machine
  Learning-Volume 70}, 1126--1135. JMLR. org.

\bibitem[{Fr{\'e}nay and Verleysen(2013)}]{frenay2013classification}
Fr{\'e}nay, B.; and Verleysen, M. 2013.
\newblock Classification in the presence of label noise: a survey.
\newblock \emph{IEEE transactions on neural networks and learning systems}
  25(5): 845--869.

\bibitem[{Goldberger and Ben{-}Reuven(2017)}]{goldberger2016training}
Goldberger, J.; and Ben{-}Reuven, E. 2017.
\newblock Training deep neural-networks using a noise adaptation layer.
\newblock In \emph{5th International Conference on Learning Representations,
  {ICLR} 2017, Toulon, France, April 24-26, 2017, Conference Track
  Proceedings}. OpenReview.net.

\bibitem[{Han et~al.(2018{\natexlab{a}})Han, Yao, Niu, Zhou, Tsang, Zhang, and
  Sugiyama}]{han2018masking}
Han, B.; Yao, J.; Niu, G.; Zhou, M.; Tsang, I.; Zhang, Y.; and Sugiyama, M.
  2018{\natexlab{a}}.
\newblock Masking: A new perspective of noisy supervision.
\newblock \emph{Advances in Neural Information Processing Systems} 31:
  5836--5846.

\bibitem[{Han et~al.(2018{\natexlab{b}})Han, Yao, Yu, Niu, Xu, Hu, Tsang, and
  Sugiyama}]{han2018co}
Han, B.; Yao, Q.; Yu, X.; Niu, G.; Xu, M.; Hu, W.; Tsang, I.; and Sugiyama, M.
  2018{\natexlab{b}}.
\newblock Co-teaching: Robust training of deep neural networks with extremely
  noisy labels.
\newblock In \emph{Advances in neural information processing systems},
  8527--8537.

\bibitem[{He et~al.(2016)He, Zhang, Ren, and Sun}]{he2016deep}
He, K.; Zhang, X.; Ren, S.; and Sun, J. 2016.
\newblock Deep residual learning for image recognition.
\newblock In \emph{Proceedings of the IEEE conference on computer vision and
  pattern recognition}, 770--778.

\bibitem[{Hendrycks et~al.(2018)Hendrycks, Mazeika, Wilson, and
  Gimpel}]{hendrycks2018using}
Hendrycks, D.; Mazeika, M.; Wilson, D.; and Gimpel, K. 2018.
\newblock Using trusted data to train deep networks on labels corrupted by
  severe noise.
\newblock In \emph{Advances in Neural Information Processing Systems},
  10477--10486.

\bibitem[{Jiang et~al.(2017)Jiang, Zhou, Leung, Li, and
  Fei-Fei}]{jiang2017mentornet}
Jiang, L.; Zhou, Z.; Leung, T.; Li, L.-J.; and Fei-Fei, L. 2017.
\newblock Mentornet: Learning data-driven curriculum for very deep neural
  networks on corrupted labels.
\newblock \emph{arXiv preprint arXiv:1712.05055} .

\bibitem[{Kingma and Ba(2014)}]{kingma2014adam}
Kingma, D.~P.; and Ba, J. 2014.
\newblock Adam: A method for stochastic optimization.
\newblock \emph{arXiv preprint arXiv:1412.6980} .

\bibitem[{Krizhevsky(2009)}]{krizhevsky2009learning}
Krizhevsky, A. 2009.
\newblock Learning multiple layers of features from tiny images.
\newblock \emph{Master's thesis, Department of Computer Science, University of
  Toronto} .

\bibitem[{Larsen et~al.(1998)Larsen, Nonboe, Hintz-Madsen, and
  Hansen}]{larsen1998design}
Larsen, J.; Nonboe, L.; Hintz-Madsen, M.; and Hansen, L.~K. 1998.
\newblock Design of robust neural network classifiers.
\newblock In \emph{Proceedings of the 1998 IEEE International Conference on
  Acoustics, Speech and Signal Processing, ICASSP'98 (Cat. No. 98CH36181)},
  volume~2, 1205--1208. IEEE.

\bibitem[{Li et~al.(2019)Li, Wong, Zhao, and Kankanhalli}]{li2019learning}
Li, J.; Wong, Y.; Zhao, Q.; and Kankanhalli, M.~S. 2019.
\newblock Learning to learn from noisy labeled data.
\newblock In \emph{Proceedings of the IEEE Conference on Computer Vision and
  Pattern Recognition}, 5051--5059.

\bibitem[{Li et~al.(2017)Li, Yang, Song, Cao, Luo, and Li}]{Li_2017_ICCV}
Li, Y.; Yang, J.; Song, Y.; Cao, L.; Luo, J.; and Li, L.-J. 2017.
\newblock Learning From Noisy Labels With Distillation.
\newblock In \emph{The IEEE International Conference on Computer Vision
  (ICCV)}.

\bibitem[{Liu, Simonyan, and Yang(2019)}]{liu2018darts}
Liu, H.; Simonyan, K.; and Yang, Y. 2019.
\newblock {DARTS:} Differentiable Architecture Search.
\newblock In \emph{7th International Conference on Learning Representations,
  {ICLR} 2019, New Orleans, LA, USA, May 6-9, 2019}. OpenReview.net.

\bibitem[{Maclaurin, Duvenaud, and Adams(2015)}]{maclaurin2015gradient}
Maclaurin, D.; Duvenaud, D.; and Adams, R. 2015.
\newblock Gradient-based hyperparameter optimization through reversible
  learning.
\newblock In \emph{International Conference on Machine Learning}, 2113--2122.

\bibitem[{Mnih and Hinton(2012)}]{mnih2012learning}
Mnih, V.; and Hinton, G.~E. 2012.
\newblock Learning to label aerial images from noisy data.
\newblock In \emph{Proceedings of the 29th International conference on machine
  learning (ICML-12)}, 567--574.

\bibitem[{Natarajan et~al.(2013)Natarajan, Dhillon, Ravikumar, and
  Tewari}]{natarajan2013learning}
Natarajan, N.; Dhillon, I.~S.; Ravikumar, P.~K.; and Tewari, A. 2013.
\newblock Learning with noisy labels.
\newblock In \emph{Advances in neural information processing systems},
  1196--1204.

\bibitem[{Nettleton, Orriols-Puig, and Fornells(2010)}]{nettleton2010study}
Nettleton, D.~F.; Orriols-Puig, A.; and Fornells, A. 2010.
\newblock A study of the effect of different types of noise on the precision of
  supervised learning techniques.
\newblock \emph{Artificial intelligence review} 33(4): 275--306.

\bibitem[{Nichol, Achiam, and Schulman(2018)}]{nichol2018first}
Nichol, A.; Achiam, J.; and Schulman, J. 2018.
\newblock On first-order meta-learning algorithms.
\newblock \emph{arXiv preprint arXiv:1803.02999} .

\bibitem[{Patrini et~al.(2017)Patrini, Rozza, Krishna~Menon, Nock, and
  Qu}]{patrini2017making}
Patrini, G.; Rozza, A.; Krishna~Menon, A.; Nock, R.; and Qu, L. 2017.
\newblock Making deep neural networks robust to label noise: A loss correction
  approach.
\newblock In \emph{Proceedings of the IEEE Conference on Computer Vision and
  Pattern Recognition}, 1944--1952.

\bibitem[{Pedregosa(2016)}]{pedregosa2016hyperparameter}
Pedregosa, F. 2016.
\newblock Hyperparameter optimization with approximate gradient.
\newblock \emph{arXiv preprint arXiv:1602.02355} .

\bibitem[{Pham et~al.(2020)Pham, Xie, Dai, and Le}]{pham2020meta}
Pham, H.; Xie, Q.; Dai, Z.; and Le, Q.~V. 2020.
\newblock Meta pseudo labels.
\newblock \emph{arXiv preprint arXiv:2003.10580} .

\bibitem[{Ranzato et~al.(2015)Ranzato, Chopra, Auli, and
  Zaremba}]{ranzato2015sequence}
Ranzato, M.; Chopra, S.; Auli, M.; and Zaremba, W. 2015.
\newblock Sequence level training with recurrent neural networks.
\newblock \emph{arXiv preprint arXiv:1511.06732} .

\bibitem[{Ravi and Larochelle(2017)}]{DBLP:conf/iclr/RaviL17}
Ravi, S.; and Larochelle, H. 2017.
\newblock Optimization as a Model for Few-Shot Learning.
\newblock In \emph{5th International Conference on Learning Representations}.

\bibitem[{Reed et~al.(2014)Reed, Lee, Anguelov, Szegedy, Erhan, and
  Rabinovich}]{reed2014training}
Reed, S.; Lee, H.; Anguelov, D.; Szegedy, C.; Erhan, D.; and Rabinovich, A.
  2014.
\newblock Training deep neural networks on noisy labels with bootstrapping.
\newblock \emph{arXiv preprint arXiv:1412.6596} .

\bibitem[{Ren et~al.(2018)Ren, Zeng, Yang, and Urtasun}]{ren2018learning}
Ren, M.; Zeng, W.; Yang, B.; and Urtasun, R. 2018.
\newblock Learning to Reweight Examples for Robust Deep Learning.
\newblock In \emph{International Conference on Machine Learning}, 4334--4343.

\bibitem[{Shu et~al.(2019)Shu, Xie, Yi, Zhao, Zhou, Xu, and Meng}]{shu2019meta}
Shu, J.; Xie, Q.; Yi, L.; Zhao, Q.; Zhou, S.; Xu, Z.; and Meng, D. 2019.
\newblock Meta-weight-net: Learning an explicit mapping for sample weighting.
\newblock In \emph{Advances in Neural Information Processing Systems},
  1917--1928.

\bibitem[{Sukhbaatar et~al.(2014)Sukhbaatar, Bruna, Paluri, Bourdev, and
  Fergus}]{sukhbaatar2014training}
Sukhbaatar, S.; Bruna, J.; Paluri, M.; Bourdev, L.; and Fergus, R. 2014.
\newblock Training convolutional networks with noisy labels.
\newblock \emph{arXiv preprint arXiv:1406.2080} .

\bibitem[{Tanaka et~al.(2018)Tanaka, Ikami, Yamasaki, and
  Aizawa}]{tanaka2018joint}
Tanaka, D.; Ikami, D.; Yamasaki, T.; and Aizawa, K. 2018.
\newblock Joint optimization framework for learning with noisy labels.
\newblock In \emph{Proceedings of the IEEE Conference on Computer Vision and
  Pattern Recognition}, 5552--5560.

\bibitem[{Veit et~al.(2017)Veit, Alldrin, Chechik, Krasin, Gupta, and
  Belongie}]{veit2017learning}
Veit, A.; Alldrin, N.; Chechik, G.; Krasin, I.; Gupta, A.; and Belongie, S.
  2017.
\newblock Learning from noisy large-scale datasets with minimal supervision.
\newblock In \emph{Proceedings of the IEEE Conference on Computer Vision and
  Pattern Recognition}, 839--847.

\bibitem[{Xia et~al.(2020)Xia, Liu, Han, Wang, Gong, Liu, Niu, Tao, and
  Sugiyama}]{xia2020part}
Xia, X.; Liu, T.; Han, B.; Wang, N.; Gong, M.; Liu, H.; Niu, G.; Tao, D.; and
  Sugiyama, M. 2020.
\newblock Part-dependent label noise: Towards instance-dependent label noise.
\newblock \emph{Advances in Neural Information Processing Systems} 33.

\bibitem[{Xia et~al.(2019)Xia, Liu, Wang, Han, Gong, Niu, and
  Sugiyama}]{xia2019anchor}
Xia, X.; Liu, T.; Wang, N.; Han, B.; Gong, C.; Niu, G.; and Sugiyama, M. 2019.
\newblock Are Anchor Points Really Indispensable in Label-Noise Learning?
\newblock In \emph{Advances in Neural Information Processing Systems},
  6838--6849.

\bibitem[{Xiao et~al.(2015)Xiao, Xia, Yang, Huang, and Wang}]{xiao2015learning}
Xiao, T.; Xia, T.; Yang, Y.; Huang, C.; and Wang, X. 2015.
\newblock Learning from massive noisy labeled data for image classification.
\newblock In \emph{Proceedings of the IEEE conference on computer vision and
  pattern recognition}, 2691--2699.

\bibitem[{Xie et~al.(2019)Xie, Dai, Hovy, Luong, and Le}]{xie2019unsupervised}
Xie, Q.; Dai, Z.; Hovy, E.; Luong, M.-T.; and Le, Q.~V. 2019.
\newblock Unsupervised data augmentation for consistency training.
\newblock \emph{arXiv preprint arXiv:1904.12848} .

\bibitem[{Yang et~al.(2016)Yang, Yang, Dyer, He, Smola, and
  Hovy}]{yang2016hierarchical}
Yang, Z.; Yang, D.; Dyer, C.; He, X.; Smola, A.; and Hovy, E. 2016.
\newblock Hierarchical attention networks for document classification.
\newblock In \emph{Proceedings of the 2016 conference of the North American
  chapter of the association for computational linguistics: human language
  technologies}, 1480--1489.

\bibitem[{Yao et~al.(2020)Yao, Liu, Han, Gong, Deng, Niu, and
  Sugiyama}]{yao2020dual}
Yao, Y.; Liu, T.; Han, B.; Gong, M.; Deng, J.; Niu, G.; and Sugiyama, M. 2020.
\newblock Dual T: Reducing estimation error for transition matrix in
  label-noise learning.
\newblock \emph{Advances in Neural Information Processing Systems} 33.

\bibitem[{Yu et~al.(2019)Yu, Han, Yao, Niu, Tsang, and
  Sugiyama}]{pmlr-v97-yu19b}
Yu, X.; Han, B.; Yao, J.; Niu, G.; Tsang, I.; and Sugiyama, M. 2019.
\newblock How does Disagreement Help Generalization against Label Corruption?
\newblock In \emph{Proceedings of the 36th International Conference on Machine
  Learning}, 7164--7173.

\bibitem[{Zhang et~al.(2017)Zhang, Bengio, Hardt, Recht, and
  Vinyals}]{zhang2016understanding}
Zhang, C.; Bengio, S.; Hardt, M.; Recht, B.; and Vinyals, O. 2017.
\newblock Understanding deep learning requires rethinking generalization.
\newblock In \emph{5th International Conference on Learning Representations,
  {ICLR} 2017, Toulon, France, April 24-26, 2017, Conference Track
  Proceedings}. OpenReview.net.

\bibitem[{Zhang, Zhao, and LeCun(2015)}]{zhang2015character}
Zhang, X.; Zhao, J.; and LeCun, Y. 2015.
\newblock Character-level convolutional networks for text classification.
\newblock In \emph{Advances in neural information processing systems},
  649--657.

\bibitem[{Zhou(2017)}]{zhou2017brief}
Zhou, Z.-H. 2017.
\newblock A brief introduction to weakly supervised learning.
\newblock \emph{National Science Review} 5(1): 44--53.

\end{thebibliography}
\end{document}